\documentclass[journal]{elsarticle}
\usepackage{textcomp}
\usepackage{adjustbox}
\usepackage{siunitx}
\usepackage{amsmath,amssymb}
\usepackage{lineno,hyperref}
\usepackage{amsmath}
\usepackage{numprint}
\usepackage{amsfonts}
\usepackage{url}
\usepackage{subfigure}
\usepackage{caption}
\usepackage{multirow}
\usepackage{breqn}
\usepackage{lipsum}
\usepackage{booktabs}
\usepackage{verbatim}
\usepackage{mathtools}
\usepackage{xcolor} 
\usepackage{graphicx} 
\usepackage{float} 
\usepackage{booktabs}
\usepackage{hyperref}
\usepackage[colorinlistoftodos,prependcaption,textsize=tiny]{todonotes}
\usepackage{upgreek}
\usepackage{nicefrac} 
\usepackage{bm}
\usepackage[algo2e,ruled,vlined]{algorithm2e}

\newcommand{\Data}{\mathcal{D}}
\newcommand{\img}{\Upomega}
\newcommand{\entr}{\mathcal{H}}
\newcommand{\paramCot}{\lambda_{\mr{cot}}}
\newcommand{\paramDiv}{\lambda_{\mr{div}}}

\newcommand{\paramMax}{\lambda_{\mr{max}}}

\newcommand{\epochIni}{t_{\mr{ini}}}
\newcommand{\epochEnd}{t_{\mr{end}}}

\newcommand{\anwer}[1]{{}}

\journal{Journal of \LaTeX\ Templates}

\DeclareMathOperator*{\argmax}{arg\,max}

\newcommand{\beq}{\begin{equation}}
\newcommand{\eeq}{\end{equation}}

\newcommand{\mr}[1]{\mathrm{#1}}
\newcommand{\tx}[1]{\textrm{#1}}

\newcommand{\Expect}{\mathbb{E}}

\newcommand{\Real}{\mathbb{R}}

\newcommand{\Classes}{\mathcal{C}}
\newcommand{\Features}{\mathcal{F}}

\newcommand{\KLdiv}{D_{\mr{KL}}}

\newcommand{\Loss}{\mathcal{L}}
\newcommand{\LossSup}{\Loss_{\mr{sup}}}
\newcommand{\LossCot}{\Loss_{\mr{cot}}}
\newcommand{\LossDiv}{\Loss_{\mr{div}}}

\newcommand{\mpred}{\overline{f}}

\newcommand{\DataLab}{\mathcal{S}}
\newcommand{\DataUnlab}{\mathcal{U}}

\begin{document}

\begin{frontmatter}

\title{Deep Co-Training for Semi-Supervised Image Segmentation}

\tnotetext[mytitlenote]{This research was funded by the Natural Sciences and Engineering Research Council of Canada.}

\author[add1]{Jizong Peng\corref{cor1}}
\ead{jizong.peng.1@etsmtl.net}
\author[add2]{Guillermo Estrada}
\ead{guillermo@ele.puc-rio.br}
\author[add1]{Marco Pedersoli}
\ead{marco.pedersoli@etsmtl.ca}
\author[add1]{Christian Desrosiers}
\ead{christian.desrosiers@etsmtl.ca}

\cortext[cor1]{Corresponding author}


\address[add1]{ETS Montreal, 1100 Notre-Dame W., Montreal, Canada}
\address[add2]{PUC-Rio, 225 Marqu\^es de S\~ao Vicente Street, Rio de Janeiro, Brazil}

\begin{abstract}
In this paper, we aim to improve the performance of semantic image segmentation in a semi-supervised setting where training is performed with a reduced set of annotated images and additional non-annotated images. We present a method based on an ensemble of deep segmentation models. Models are trained on subsets of the annotated data and use non-annotated images to exchange information with each other, similar to co-training. Diversity across models is enforced with the use of adversarial samples. We demonstrate the potential of our method on two challenging image segmentation problems, and illustrate its ability to share information between simultaneously trained models, while preserving their diversity. Results indicate clear advantages in terms of performance compared to recently proposed semi-supervised methods for segmentation.   
\end{abstract}

\begin{keyword}
Deep learning \sep semi-supervised learning \sep ensemble learning \sep co-training \sep image segmentation   
\end{keyword}

\end{frontmatter}


\section{Introduction}

Semantic segmentation \cite{noh2015learning} is a fundamental problem in computer vision, which requires assigning the proper category label to each pixel of a given image. It plays a key role in applications of various domains, including image retrieval, autonomous driving, video surveillance, remote sensing, robotics and biomedical imaging. This task is particularly important for medical image analysis, where it serves as a necessary pre-processing step for the assessment and treatment planning of various medical conditions \cite{Litjens2017}.

In recent years, supervised approaches, in particular those based on deep learning, have shown tremendous potential for automated image segmentation. In such approaches, parametric models like fully-convolutional neural networks (F-CNNs) \cite{long2015fully} are trained with a large set of annotated images by minimizing some loss function like cross-entropy or Dice loss \cite{milletari2016v}. In many cases, however, obtaining sufficient data for training can be challenging, and manually annotating images can be a time consuming task \cite{kolesnikov2016seed}. This problem is even more significant in medical imaging applications, where images are typically 3D volumes (e.g., MRI or CT scans), the regions to delineate have low contrast, and annotations must be made by highly-trained experts. For challenging problems like infant brain segmentation, obtaining reliable annotations for a single subject may take a radiologist up to a week\footnote{See \url{http://iseg2017.web.unc.edu/reference/}} \cite{wang2019benchmark}.

To alleviate the need for fully-annotated data, numerous works have focused on developing weakly-supervised methods for segmentation. In such methods, easier to obtain annotations like image-level tags \cite{pinheiro2015weakly,papandreou2015weakly,kervadec2019constrained,pathak2015constrained}, bounding boxes \cite{dai2015boxsup,rajchl2017deepcut} or scribbles \cite{lin2016scribblesup} are used for training segmentation models, instead of whole-image pixel labels. Multiple instance learning (MIL) \cite{vezhnevets2010towards} is a popular technique for dealing with image tags, where images are considered as bags of pixels\,/\,superpixels (i.e., instances) and positive examples for a given object of interest (i.e., tag) are images for which at least one pixel\,/\,superpixel corresponds to that object. MIL methods for segmentation typically rely on objectness \cite{bearman2016s,wei2016learning,pinheiro2015image,qi2016augmented,saleh2016built,shimoda2016distinct}, class-specific saliency and activation maps \cite{hou2017deeply,liu2016dhsnet,selvaraju2017grad,zhou2016learning}, or image-level constraints \cite{kervadec2019constrained,pathak2015constrained} to obtain a prior on the presence or location of objects in the image. 

In various scenarios, weakly-supervised learning methods for segmentation may not be suitable. For instance, adding bounding boxes or point annotations can still be time-costly for 3D scans, which may contain over 100 separate images (i.e., 2D slices). Likewise, image-level tags may not be useful in segmentation tasks where one must separate a single region of interest (i.e., foreground) from the background. In contrast, semi-supervised learning methods \cite{bai2017semi,baur2017semi,min2018robust,zhou2018semi,perone2018unsupervised} seek to improve the training of segmentation models by leveraging unlabeled images, in addition to labeled ones. Unlike weakly-supervised approaches, these methods rely on intrinsic properties of the data distribution (or \emph{priors}) which are not specific to individual images. Semi-supervised methods for segmentation include techniques based on self-training \cite{bai2017semi}, model-based \cite{gupta2016cross} or data-based \cite{radosavovic2017data,zhou2018semi} distillation 
, attention learning \cite{min2018robust}, adversarial learning \cite{souly2017semi,Hung_semiseg_2018,zhang2017deep,luc2016semantic}, and manifold embedding \cite{baur2017semi}. 

Co-training is one of the most popular general-purpose techniques for semi-supervised learning. 
This technique originally proposed by Blum and Mitchell \cite{blum1998combining} is based on the idea that training examples can be described by two complementary (conditionally independent given the corresponding class labels) sets of features, called views. 
Multi-view learning \cite{xu2013survey} extends this idea to multiple complementary views. The general principle of this type of method is to simultaneously train classifiers for each view, using the labeled data, such that their predictions agree for unlabeled examples. Enforcing this agreement between classifiers reduces the search space and thus helps find a model which will generalize well to unseen data. While co-training and learning methods 
have been used with great success in natural language processing \cite{wan2009co,nigam2000understanding,maeireizo2004co}, their application to visual tasks has so far been limited \cite{levin2003unsupervised}. One of the main reasons for this is that such methods require complementary models to learn from independent features. Although such independent features may be available in specific scenarios (e.g., multiplanar images \cite{zhou2018semi}), there is no effective way to construct these sets from individual images.  
Recently, Qiao \emph{et al.} proposed a deep co-training method for semi-supervised image recognition \cite{qiao2018deep}. The main innovation of this work is to use adversarial examples, built from both labeled and unlabeled images, for imposing diversity among the different classifiers. Specifically, during training, a classifier is encouraged to output predictions similar to those of the other classifier for adversarial examples, hence classifiers will tend to disagree for those examples. 

Until now, deep co-training has been applied only to classification. In contrast, semantic segmentation is a more complex problem with a larger and structured output space. In this work we extend and adapt the co-training approach for this task. The contributions of our work are as follows:
\begin{itemize}
    \item We present a deep adversarial co-training method for semantic segmentation, extending the work of Qiao \emph{et al.} to this more challenging problem. To our knowledge, this is the first co-training method proposed for single-image semantic segmentation.
    
    \item We show key differences between the application of deep co-training for classification and segmentation, and explore the effect of adversarial training on the prediction diversity of segmentation models.
    
    \item We conduct a comprehensive set of experiments which demonstrate the potential of co-training for segmenting different types of images. Our experiments also analyze the impact of various elements of the method, including the number of classifiers, the trade-off between model agreement and diversity, and the generation of adversarial examples. We believe these experiments can be of benefit to future investigations on co-training methods for segmentation.  
\end{itemize}

The rest of this paper is as follows. In the next section, we give a brief summary of related literature, focusing on recently proposed methods for semi-supervised segmentation. In Section \ref{sec:method}, we present our deep adversarial co-training approach for segmentation. We then evaluate our method on the tasks of segmenting cardiac and spine structures in section \ref{sec:experiments}. Finally, we conclude with a summary of our contribution and results.

\section{Related work}\label{sec:related-work}

Semi-supervised learning has a long history in machine learning. The first methods were proposed around 50 years ago for estimating mixture models \cite{Cooper1970On,Dempster1977Maximum}. Since then, many different approaches have been proposed. Here, we will focus mostly on the most recent and promising methods for visual recognition and, more specifically, semantic segmentation. For a complete review of semi-supervised methods, see \cite{Chapelle2010Semi}.

A quite simple, yet powerful approach for semi-supervised learning is to select the most likely label of the current model as ground truth for unsupervised data. This is often referred to as pseudo-label \cite{Leepseudolabel} or entropy regularization \cite{Grandvalet+Bengio-ssl-2006}.
More sophisticated approaches make use of unlabeled samples, leveraging the unsupervised representation of an autoencoder \cite{Ramus15Semi} or a variational autoencoder \cite{Kingma14Semi}.
Another line of research for semi-supervised learning is based on the idea that the pseudo-labeling can be improved and made more robust if multiple models are used for generating the pseudo-labels \cite{Laine16Temporal,Tarvainen17Weight}.
Regularizing the learning with adversarial examples is also a promising technique. It consists in generating samples that are adversarial to the model \cite{Goodfellow15Explaining}, i.e. samples that the model cannot classify correctly, and adding them to the training data to improve robustness.
Recently, the generation of adversarial samples has been applied to unlabeled samples, therefore extending their use to semi-supervised learning with very promising results \cite{Miyato15Virtual}. This technique has also been used for co-training multiple classification models \cite{qiao2018deep}. Our proposed method is based on the last approach, but adapted to the more challenging task of semi-supervised image segmentation. 
For an updated evaluation of state-of-the art semi-supervised methods for image classification, see \cite{Oliver2018Realistic}.

Semi-supervised learning has also been used for image segmentation \cite{bai2017semi,baur2017semi,min2018robust,zhou2018semi}. As for classification, the main idea of semi-supervised segmentation methods is to propagate the labels of training samples to unlabeled images. However, in the case of segmentation, the output is structured and therefore methods based on local vicinity of the sample representation would not work. 
A common approach is to use an iterative two steps procedure in which: \emph{i}) the unlabeled images are annotated considering the output of the segmentation network as ground truth; \emph{ii}) the network parameters are updated based on the segmented (annotated) images \cite{bai2017semi}. A common problem of such approach is that initial small errors might be propagated and amplified to unlabeled images, producing catastrophic results. Various approaches are used to avoid this problem. For instance, model-based \cite{gupta2016cross} and data-based \cite{radosavovic2017data,zhou2018semi} distillation can reduce the error propagation by aggregating the prediction of multiple teacher models to train a student model \cite{min2018robust}. 
Another approach proposed by Baur \emph{et al.} \cite{baur2017semi} embeds the network representation in a manifold, such that images having similar characteristics are near to each other. 

Methods based on generative adversarial networks (GANs) \cite{Goodfellow2014GAN} have recently shown promising results for semi-supervised segmentation \cite{souly2017semi,Hung_semiseg_2018,zhang2017deep,luc2016semantic}. 
The first approach using GANs for semantic segmentation was proposed by Luc \emph{et al.} \cite{luc2016semantic} and extended to the semi-supervised case in \cite{zhang2017deep}. In this work, a discriminator network should distinguish between the segmentation of labeled and unlabeled images. This forces the segmentation model to perform as well on unlabeled images, in order to fool the discriminator. An improved strategy is proposed by Hung \emph{et al.} \cite{Hung_semiseg_2018}, where the discriminator is used to predict areas of high confidence on unlabeled images. These areas are then used to update the segmentation network. It is important to distinguish GAN models from the use of adversarial examples \cite{Goodfellow15Explaining}. While GAN models are based on the simultaneous learning of two adversarial networks (the discriminator and the generator), adversarial training proposes the generation of samples with subtle modifications that can fool a learned model.
Although GANs have already been employed for improving semi-supervised approaches, adversarial samples have not yet been applied to segmentation. In this paper, we show how to leverage adversarial samples in semi-supervised segmentation by exploiting a co-training procedure \cite{qiao2018deep}.

\begin{figure}[t!]
\centering
\includegraphics[width=0.8\textwidth]{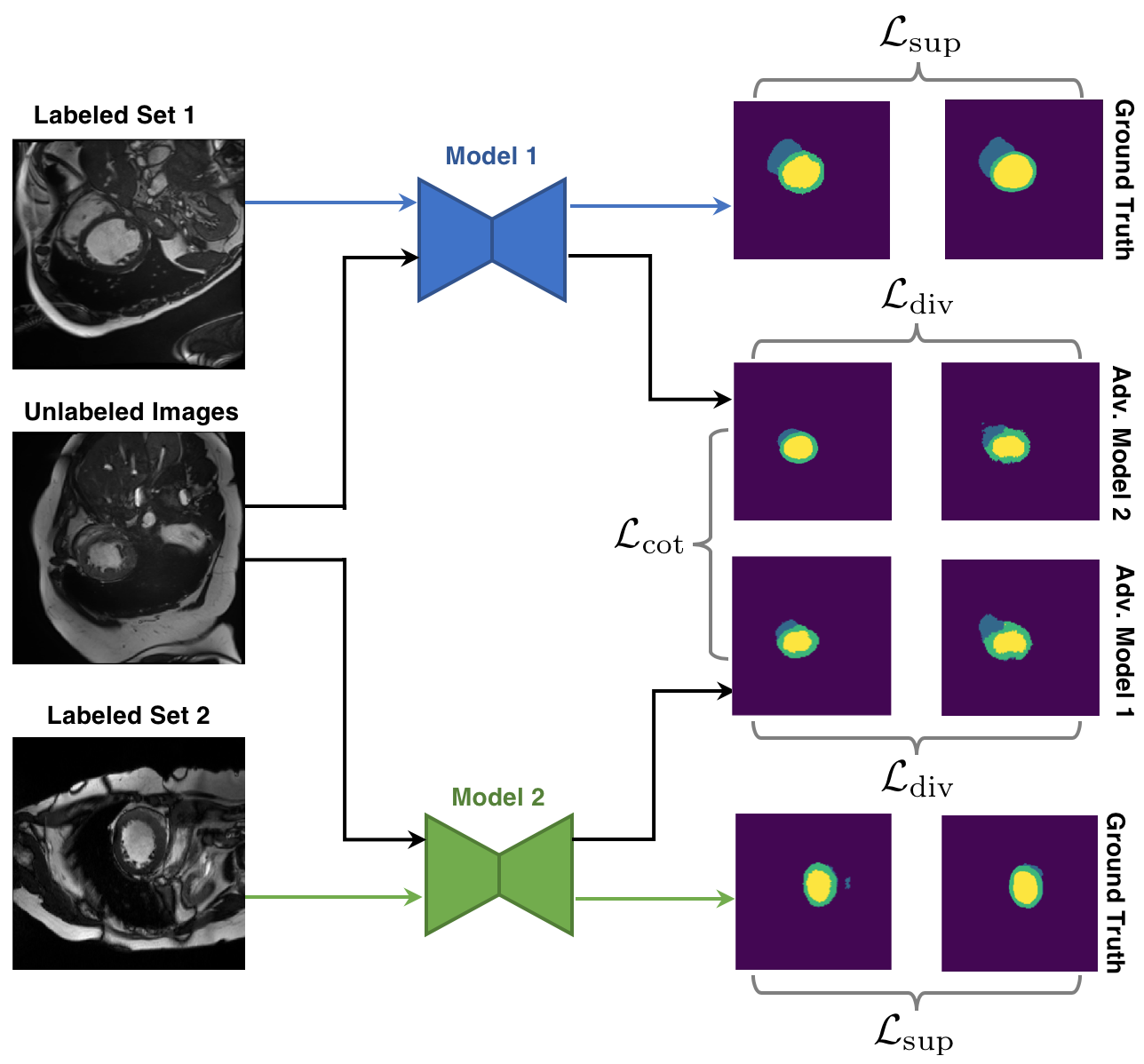}
\caption{Overview of the deep co-training approach proposed for image segmentation (dual-view setting). Two deep CNN models are trained simultaneously with different sets of labeled images and a common set of unlabeled images. The loss function is composed of three terms: $\LossSup$, $\LossCot$ and $\LossDiv$. Term $\LossSup$ ensures that network predictions for labeled examples are consistent with ground truth segmentation masks; $\LossCot$ forces networks to agree with each other for unlabeled examples; $\LossDiv$ imposes a network to agree with the predictions of the other network's adversarial examples.}
\label{fig:co-training-diagram}
\end{figure}

\section{Methodology}\label{sec:method}

\subsection{Problem formulation}

As a dense prediction problem with complex output space, semantic segmentation is extremely challenging in a semi-supervised setting. 
In real-life applications, particularly those related to medical imaging, such a setting is however common since manual annotation is often an expensive and time-consuming process. Consequently, only a small fraction of images in the dataset can have full pixel-wise labels. The proposed method aims to exploit both labeled and unlabeled images by using the general, yet powerful principle of multi-view co-training. 

We formalize the problem of image segmentation as follows. Given a set of labeled data $\DataLab = \{(x_1,y_1), \ldots, (x_{m},y_{m})\}$, each example comprised of an image $x_i : \img \to \Features$ and corresponding ground truth segmentation mask $y : \Omega \to \Classes$, where $\Omega$ is the set of image pixels (or voxels in the 3D case), $\Features$ the set of pixel features (e.g., $\Features = \Real$ for grey-scale images), and $\Classes$ the set of possible labels. In a semi-supervised setting, we also have a set of $n$ unlabeled images $\DataUnlab = \{x_1, \ldots, x_{n}\}$, with $n \gg m$, without ground truth labels. The goal is to learn from $\Data = \DataLab \cup \DataUnlab$ a segmentation model $f$ parametrized by $\theta$, which maps each pixel of an input image to its correct label.

\subsection{Proposed approach}

As in standard multi-view learning approaches, we train multiple models in a collaborative manner and, once trained, combine their outputs to predict the labels of new images. Motivated by the outstanding performance of deep convolutional network networks (CNNs) for various segmentation tasks \cite{litjens2017survey,dolz20173d,dolz2018hyperdense}, we employ this type of model in the proposed approach. Specifically, we train an ensemble of $k$ segmentation networks $f^i(\cdot \,; \,\theta^i)$, $i=1, \ldots, k$. 
We assume the network uses a softmax function at each image pixel to compute label probabilities, 
and denote as $f^i_{jc}$ the probability of label $c$ for pixel $j$, predicted by model $i$. Without loss of generality, in what follows, we will consider a dual view setting (i.e., $k=2$) and describe how this setting can be naturally extended to multiple views.

Following co-training methods for classification,
we employ a loss function composed of a weighted sum of three separate terms to train the ensemble's segmentation models (see Fig. \ref{fig:co-training-diagram}):
\beq
 \Loss(\theta; \, \Data) \ = \ \LossSup(\theta; \, \DataLab) \ + \ \paramCot \, \LossCot(\theta; \, \DataUnlab) \ + \ \paramDiv \, \LossDiv(\theta; \, \Data).
 \label{equ:obj}
\eeq
The three loss terms are explained in following subsections.


\subsubsection{Supervised loss}

The first term, $\LossSup$, is the supervised loss obtained from labeled examples. It aggregates the loss computed separately for each model:  
\beq\label{eq:loss-sup}
    \LossSup(\theta; \, \DataLab) \ = \ \LossSup^1(\theta^1; \, \DataLab^1) + \LossSup^2(\theta^2; \, \DataLab^2).
\eeq
Here, labeled data subsets $\DataLab^i \subset \DataLab$, $i \in \{1,2\}$ can differ across models to ensure their diversity. While any segmentation loss can be considered, in this work, we employed the well-known pixel-wise cross-entropy loss, defined as
\beq
\LossSup^i(\theta^i; \, \DataLab^i) \ = \ \Expect_{(x,y)\in \DataLab^i} \left[\sum_{j\in \img} \sum_{c \in \Classes} y_{jc} \log f^i_{jc}(x; \theta^i)\right],
\eeq 
where $y_{jc}=1$ if the true label of pixel $j$ is $c$, else $y_{jc}=0$ (i.e., one-hot label encoding).
Supervised loss $\LossSup$ encourages models to output consistent predictions with respect to their ground truth labels.

\subsubsection{Ensemble agreement loss}

In addition to exploiting labeled information, unlabeled image dataset $\DataUnlab$ is also used to guide the learning process. Based on the consensus principle \cite{xu2013survey}, we want the segmentation networks to output similar predictions for the same unlabeled images. We argue that enforcing this agreement helps improve the generalization of individual models by restricting their parameter search space to cross-view consistent solutions. Toward this goal, we minimize the distance between the class distributions predicted by different models. To make our approach compatible with more than two views, we define the agreement loss $\LossCot$ as the Jensen-Shannon divergence (JSD), which is the average Kullack-Liebler divergence $\KLdiv$ between the prediction of each model $f^i$ and their mean prediction $\mpred$:
\begin{align}
  & \LossCot(\theta; \, \DataUnlab) \ = \ \Expect_{x \in \DataUnlab} \left[
    \KLdiv\Big(f^1(x; \, \theta^1) \ || \ \mpred(x; \, \theta)\Big) \, + \, \KLdiv\Big(f^2(x; \, \theta^2) \ || \ \mpred(x; \, \theta)\Big)\right] \nonumber \\
    & \ \ = \ \Expect_{x \in \DataUnlab} \left[ \entr\Big(\tfrac{1}{2}\big(f^1(x; \, \theta^1) + f^2(x; \, \theta^2)\big)\Big) \, - \, 
\frac{1}{2} \Big(\entr\big(f^1(x; \, \theta^1)\big) \, + \, \entr\big(f^2(x; \, \theta^2)\big)\Big)\right].
\end{align}
In this equation, $\entr(\cdot)$ corresponds to the Shannon entropy. Unlike KL divergence, the JSD between different distributions is symmetric, and thus loss $\LossCot$ considers the prediction of all models equally important when minimizing their disagreement.

\begin{figure}
\centering
\includegraphics[width=1\textwidth]{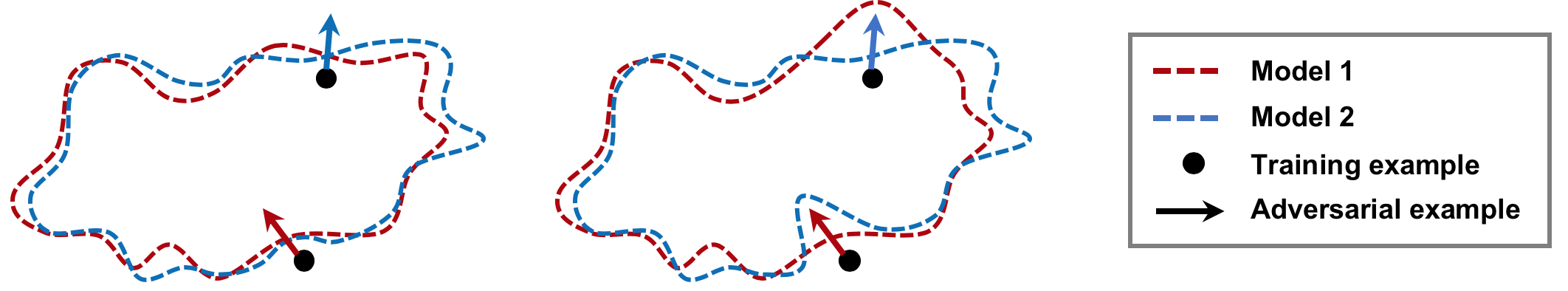}
\caption{Illustration of the ensemble diversity strategy based on adversarial training. Adversarial examples are generated from training images (black dots), for both models (red and blue arrows). Each model is then forced to agree with the prediction of the other model for its own adversarial examples (right-side image).}
\label{fig:co-training-adv}
\end{figure}

\subsubsection{Diversity loss}

A key principle of ensemble learning is having diversity between models in the ensemble. If all models learn the same class distribution, then combining their output will not be superior to individual model predictions. In co-training, diversity is essential so that models can learn from one another during training. The standard approach for obtaining diversity is to have independent sets of features (i.e., views), or generating them by splitting available features into complementary subsets. In deep CNN classification, however, the internal representation of images is learned by the network during training, therefore such standard approach cannot be applied. Instead, we define diversity based on network output, and consider two models as different if they predict sufficiently different segmentations for some given images.

Since models in the ensemble must agree for unlabeled images, and their prediction on labeled images is constrained by ground-truth segmentation masks, training images cannot be used directly to impose diversity. Instead, we use the approach proposed by Qiao \emph{et al.} for image classification \cite{qiao2018deep}, and augment the dataset with adversarial examples generated from both labeled and unlabeled data. 
Adversarial examples for a model are used to teach other models in the ensemble. In the case of dual-view co-training, we define our diversity loss as 
\beq\label{eq:jadv}
  \LossDiv(\theta; \Data) \, = \, \Expect_{x \in \Data} \left[ \entr\Big(f^1\big(x; \theta^1\big), \, f^2\big(g^1(x); \theta^2\big)\Big) \, + \, 
    \entr\Big(f^2\big(x; \theta^2\big), \, f^1\big(g^2(x); \theta^1\big)\Big)\right],
\eeq
where $\entr(\cdot,\cdot)$ refers to cross-entropy and $g^i(x)$ is an adversarial example targeted on model $f^i(\cdot; \theta^i)$, given input image $x$. As illustrated in Fig. \ref{fig:co-training-adv}, this loss function encourages a model to be robust to the adversarial examples generated for 
the other one, thereby avoiding the collapse of their decision boundary on each other (i.e., the adversarial loss reaches its maximum value when the two networks are identical). 

The diversity imposed by the loss can also be motivated as follows. If example $g^1(x)$ is adversarial for model 1, then we have that $f^1\big(x; \, \theta^1\big) \neq f^1\big(g^1(x); \, \theta^1\big)$. Moreover, minimizing the first term of Eq. (\ref{eq:jadv}) will impose that $f^1\big(x; \, \theta^1\big) = f^2\big(g^1(x); \, \theta^2\big)$. Last, combining both relations yields $f^1\big(g^1(x); \, \theta^1\big) \neq f^2\big(g^1(x); \, \theta^2\big)$. Applying the same idea for model 2, we conclude that models will disagree on adversarial examples of each model. One should note, however, that the above relations are not guaranteed to hold in practice (e.g., predictions can be very similar but not equal). In our experiments, we show that differences mostly occur on the boundary between different regions, which is where most segmentation mistakes are made (see Fig. \ref{fig:model-disagreement}).

Adversarial examples are generated by adding small perturbations to input images, so as to change the network's 
prediction as much as possible. In this work, we generate these examples using distinct schemes depending on the source of the image $x$. If $x$ is drawn from the unlabeled dataset $\DataUnlab$, we apply the Virtual Adversarial Training (VAT) \cite{miyato2018virtual} method because no ground truth is available. VAT optimizes local distribution smoothness (LDS) which measures the robustness of the model against virtual adversarial direction. Following VAT, we generate an adversarial example from training image $x$ as $x_{\mr{adv}} = x + r_{\mr{adv}}$, where
\beq
    r_{\mr{adv}} \ = \ \argmax_{r; \, \|r\|_2 \leq \epsilon}
        \ \KLdiv\big(f(x; \, \theta) \ || \ f(x + r; \, \theta)\big).
\eeq
On the other hand, when $x$ is drawn from the labeled set $\DataLab$, we instead apply the Fast Gradient Sign Method (FGSM) since it can produce noise targeted to the ground truth, thus providing more valuable information. In this case, adversarial examples $x_{\mr{adv}}$ are generated with FGSM as
\beq
 x_{\mr{adv}} \ = \ x \, + \, \epsilon \cdot 
 \mr{sign}\Big(\nabla_x \entr\big(f(x; \, \theta), y\big)\Big),
\eeq
where $\entr$ is the cross-entropy loss used as in full supervision
, and $y$ is the true label of $x$. This approach also constrains the magnitude of adversarial perturbations using a predefined $\epsilon$ parameter. 


\begin{center}
\begin{footnotesize}
\begin{algorithm2e}[ht!]
%
\KwIn{Labeled images $\DataLab = \{(x_1,y_1), \ldots, (x_{m},y_{m})\}$;}				
\KwIn{Unlabeled images $\DataUnlab = \{x_1, \ldots, x_{n}\}$;}
\KwIn{Number of views $k$;}
\KwOut{Network parameters $\{\theta^i\}_{i=1}^k$;}
\caption{Deep Co-Training Segmentation (\emph{training})}\label{algo}

\BlankLine
Initialize network parameters $\theta^i$, $i=1,\ldots,k$\;

\BlankLine
\For{$\mr{epoch} = 1, \ldots, E_{\mr{max}}$}{
    	
    \BlankLine
    \For{$\mr{iter} = 1, \ldots, T_{\mr{max}}$}{
        \BlankLine
        Randomly choose two different networks $\theta^{i_1}$ and $\theta^{i_2}$\; 
        
        \BlankLine
        Draw two batches $\DataLab^{i_1},\, \DataLab^{i_2} \subset \DataLab$ of $b$ labeled images $(x, y)$ (with replacement)\;
        
        \BlankLine
        Draw a single batch $\DataUnlab^{b} \subset \DataUnlab$ of $b$ unlabeled images $x$\;				        				
        \BlankLine				        
        Compute adversarial examples $g^{i_1}(x)$ for all $x \in$  $\DataLab^{i_1}\,\cup\,\DataUnlab^{b}$, and $g^{i_2}(x)$ for all $x \in \DataLab^{i_2}\,\cup\,\DataUnlab^{b}$, using Eq. (6) or (7)\;
        
        \BlankLine
        Let $\Loss \ = \ \LossSup \, + \, \paramCot \, \LossCot \, + \, \paramDiv \, \LossDiv$, as defined in Eq. (\ref{eq:loss-sup})-(\ref{eq:jadv}), using $\DataLab^{i_j}$ for the supervised loss of model $i_j$\;
        
        \BlankLine
        Compute gradients w.r.t. $\Loss$ and update parameters $\theta^{i_j}$, $j=1,2$, using back-propagation\;
    }				
    
    \BlankLine
    Update learning rate and parameters $\paramCot, \paramDiv$ as in Eq. (\ref{equ:scheduler})\;
}	

\BlankLine
\Return{$\{\theta^i\}_{i=1}^k$} \;
\end{algorithm2e}
\end{footnotesize}
\end{center}

\subsubsection{Training and testing process}

\begin{figure}[!htb]
\centering
  \includegraphics[width=0.42\linewidth]{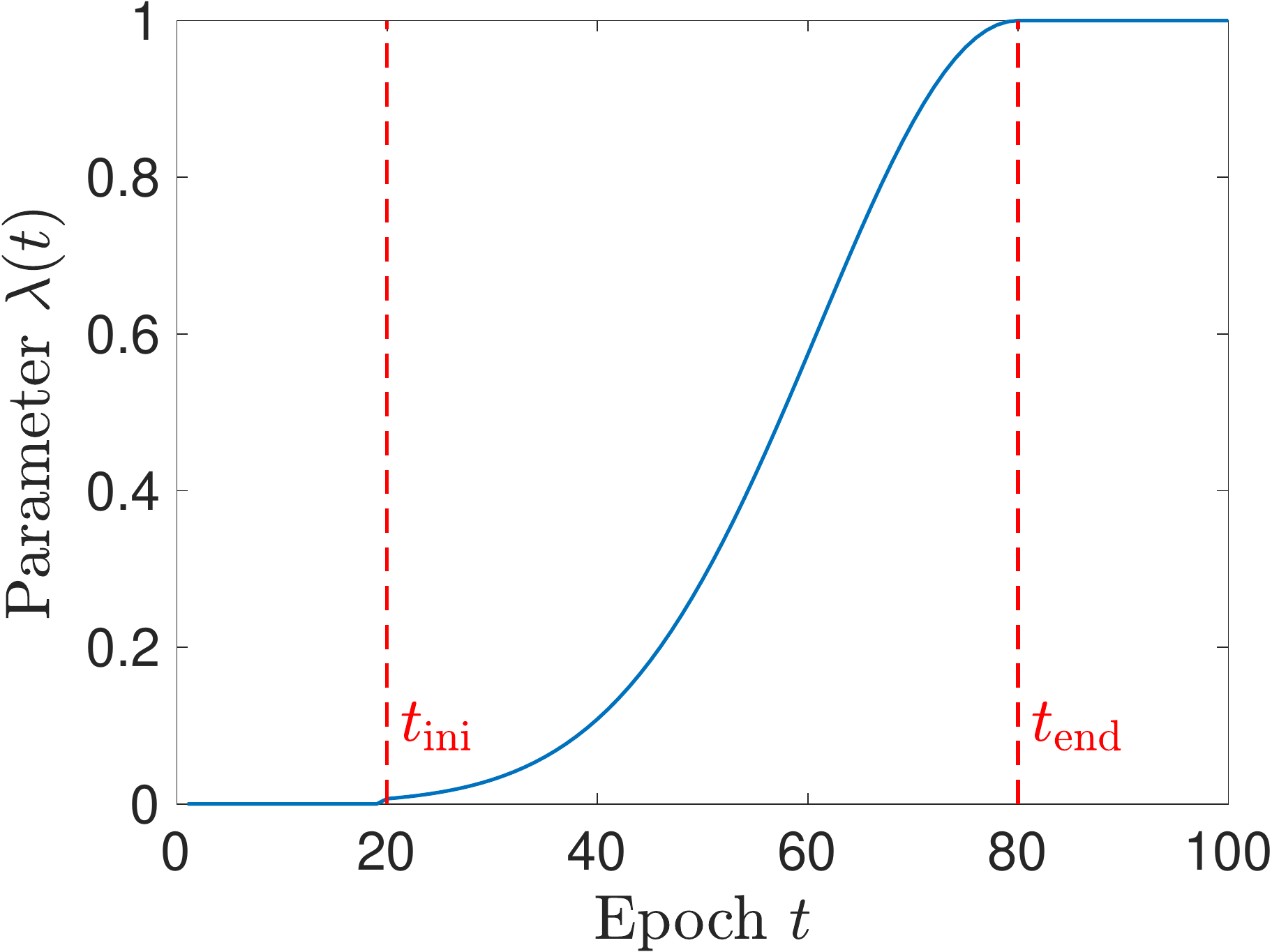}
\caption{Example of ramp-up function $\lambda(t)$ for $\paramMax=1$, $\epochIni=20$ and $\epochIni=80$.}
\label{fig:ramp-up}
\end{figure}

The whole training process is summarized in Algorithm \ref{algo}. The algorithm takes as input labeled images $\DataLab$, unlabeled images $\DataUnlab$, and the number $k$ of segmentation models to train (views). It outputs the parameters of the $k$ trained models, i.e. $\{\theta^i\}_{i=1}^k$. At every training epoch, the algorithm performs $T_{\mr{max}}$ mini-batch iterations to update the network parameters. In each iteration, we randomly select a pair of networks to generate adversarial examples and compute the supervised, co-training and diversity loss functions. In practice, network pairs are sampled such that all networks are updated at each $\lceil k/2\rceil$ iterations. At the end of each epoch, we modify the learning rate using standard decay, and update the co-training and diversity loss parameters $\paramCot$ and $\paramDiv$ with a dynamic strategy. This strategy follows a Gaussian ramp-up curve defined by parameters $\paramMax$, $\epochIni$ and $\epochEnd$:
\begin{equation}
\lambda(t) =\left\{
\begin{array}{ll}
0     & \tx{, if } \, t < \epochIni\\
\paramMax \cdot \exp\Big(-5\cdot\left(1-\frac{t \, - \, \epochIni}{\epochEnd \, - \, \epochIni}\right)^2\Big)    & \tx{, if  } \, \epochIni \leq t < \epochEnd\\
\paramMax & \tx{, if } \, t \geq \epochEnd
\end{array} \right.,
\label{equ:scheduler}
\end{equation}
An example of the ramp-up function is shown in Fig. \ref{fig:ramp-up}. The ramp-up only starts after $\epochIni$ epochs to avoid hampering training in its early stage, and reaches and its maximum value $\paramMax$ after $\epochEnd$ epochs.

In testing, we feed an unlabeled image to the trained models and combine their outputs to obtain the final segmentation. This can be done in different ways, for instance, using hard- or soft-voting. In hard-voting, the label of a pixel is the one predicted by the majority of models (with random tie-breaking). On the other hand, soft-voting consists in averaging the pixel-wise class probabilities across models, and using this average as ensemble prediction. The latter technique is commonly used in homogeneous ensemble techniques like bootstrap aggregating (bagging).

\section{Experiments and results}
\label{sec:experiments}

\subsection{Evaluation datasets and metrics}

\begin{figure}[t!]
\minipage{0.195\textwidth}
  \includegraphics[width=\linewidth]{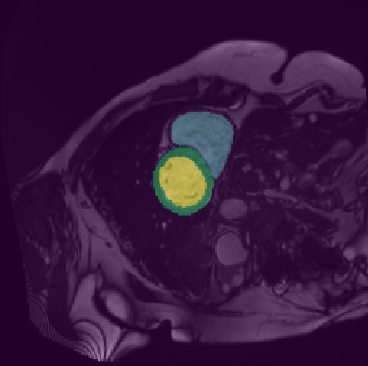}
\endminipage\hfill
\minipage{0.195\textwidth}
  \includegraphics[width=\linewidth]{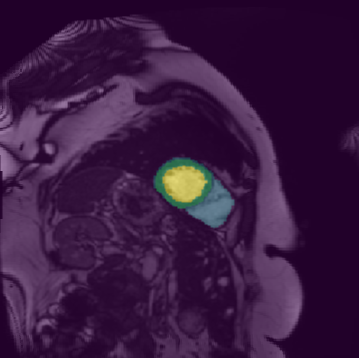}
\endminipage\hfill
\minipage{0.195\textwidth}
  \includegraphics[width=\linewidth]{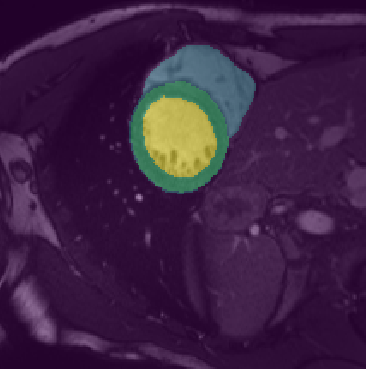}
\endminipage\hfill
\minipage{0.195\textwidth}
  \includegraphics[width=\linewidth]{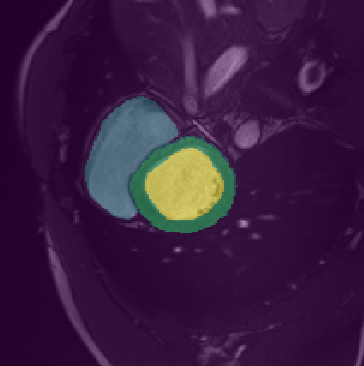}
\endminipage\hfill
\minipage{0.195\textwidth}
  \includegraphics[width=\linewidth]{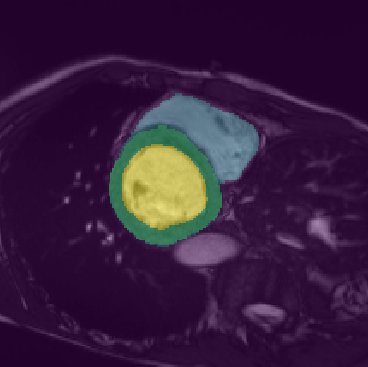}
\endminipage\hfill
\caption{Examples of images and ground truth segmentation masks in the ACDC dataset. Images are segmented in four separate classes:
endocardium of the left ventricle (LV, yellow),
myocardium of the left ventricle (Myo, green), endocardium of the right ventricle (RV, blue),
background (purple).}
\label{fig:acdc-examples}
\end{figure}

Our experiments are conducted on three clinically-relevant benchmark datasets for medical image segmentation: Automated Cardiac Diagnosis Challenge (ACDC) \cite{ACDC}, Spinal Cord Gray Matter Challenge (SCGM) \cite{SCGM}, and Spleen sub-task dataset of the Medical Segmentation Decathlon Challenge \cite{simpson2019large}.
\begin{itemize}
\item \textbf{ACDC dataset}: The publicly available ACDC dataset consists of 200 short-axis cine-MRI scans from 100 patients, evenly distributed in 5 subgroups: normal, myocardial infarction, dilated cardiomyopathy, hypertrophic cardiomyopathy, and abnormal right ventricles. Scans correspond to end-diastolic (ED) and end-systolic (ES) phases, and were acquired on 1.5T and 3T systems with resolutions ranging from 0.70\,$\times$\,0.70 mm to 1.92\,$\times$\,1.92 mm in-plane and 5 mm to 10 mm through-plane. Segmentation masks delineate 4 regions of interest: left ventricle endocardium (LV), left ventricle myocardium (Myo), right ventricle endocardium (RV), and background (see Fig. \ref{fig:acdc-examples}). For our experiments, we used a split of
75 subjects (150 scans) for training and 25 subjects (50 scans) for testing. Short-axis slices within 3D-MRI scans were considered as 2D images, which were re-sized to 256\,$\times$\,256.

\item \textbf{SCGM dataset}: The SPGM dataset is a publicly-available collection of multi-center, multi-vendor MRI. It comprises a total of 80 healthy subjects (age range of 28.3 to 44.3 years) obtained by four different centers, with 20 subjects from each center. Scans were acquired using different MRI systems and distinct acquisition parameters, leading to high-variability of image characteristics: resolution range of 0.25\,$\times$\,0.25\,$\times$\,2.5 mm to 0.5\,$\times$\,0.5\,$\times$\,5.0 mm, number of axial slices range of 3 to 28. The training set contains 40 labeled scans, each annotated slice-wise by 4 independent experts and the ground truth mask obtained by majority voting. Ground truth labels for the remaining 40 test images are not available. For additional details on the dataset, see \cite{SCGM}. 

In \cite{perone2018unsupervised}, this dataset is used to train and test a semi-supervised segmentation method based on the mean teacher algorithm. Experiments of this work, which focused on domain adaptation, used images from centers 1 and 2 as the training set, images from center 3 as the validation set, and images from center 4 as the test set. In our work, we seek to evaluate methods in a more traditional semi-supervised setting, where very few labeled images are seen in training. Hence, we consider a different training set where labeled images only come from center 1 (total of 30 images), and unlabeled images from all centers are used (total of 465 images). The test set contains labeled images from centers 3 and 4 (total of 264 images).  
Following \cite{perone2018unsupervised}, slices in each scan are first resampled to a uniform resolution of 0.25\,$\times$\,0.25 mm, and then center-cropped to a size of 200\,$\times$\,200 pixels.

\item \textbf{Spleen datset}: As one of the ten sub-tasks of the Medical Segmentation Decathlon Challenge \cite{simpson2019large}, the publicly-available Spleen dataset\footnote{\url{http://medicaldecathlon.com/}} consists of patients undergoing chemotherapy treatment for liver metastases. A total of 61 portal venous phase CT scans (only 41 were given with ground truth) were included in the dataset with acquisition and reconstruction parameters described in \cite{simpson2019large}. The ground truth segmentation was generated by a semi-automatic segmentation software and then refined by an expert abdominal radiologist.

For our experiments, 2D images are obtained by slicing the high-resolution CT volumes along the axial plane, followed by a max-min normalization with a range between 0 and 1. Each slice is then resized to a resolution of 256$\times$256 or 512$\times$512 to test the robustness of the different algorithms to various input image resolutions. In order to evaluate these algorithms in a semi-supervised setting, we split the dataset into labeled, unlabeled and validation image subsets, comprising CT scans of 4, 32, and 5 patients respectively.
\end{itemize}

As in similar studies, we use the Dice similarity coefficient (DSC) and the Hausdorff distance (HD) to evaluate the performance of segmentation models. DSC measures the overlap between the predicted segmentation $S$ and ground truth segmentation $G$:
\beq
    \mr{DSC}(S,G) \ = \ \frac{2 |S \cap G|}{\,\,|S| + |G|}.
\eeq
On the other hand, HD is a boundary distance metric which measures the largest distance (in mm) between a point in $S$ and its nearest point in $G$ (or vice-versa):
\beq
    \mr{HD}(S,G) \ = \ \max\big\{d(S,G), \, d(G,S)\big\}.
\eeq
Unlike for DSC, where a perfect segmentation has a value of $1$ and the worse possible segmentation a value of $0$, a smaller HD value indicates a better segmentation. 

\subsection{Experimental details}

As segmentation network, we employed the well-known U-Net \cite{ronneberger2015u} architecture, with 15 layers, Dropout and ReLU activations. This architecture is one of the most popular models for segmentation, especially for tasks related to medical imaging. The same data augmentation strategy was considered for all datasets, which applies random rotation, flip, and random crop of 85-95\% surface on the original image.

Networks were trained using stochastic gradient descent (SGD) with the Adam optimizer. Learning parameters were set separately for different datasets. For the ACDC and Spleen datasets, we used a maximum number of epochs of 300, an initial learning rate of 0.001 and a weight decay of 0.0001. The learning rate was decreased by a factor of 10 every 90 epochs. Batch size was set to 4 for both labeled and unlabeled data. FSGM with $\epsilon=0.03$ or VAT with $\epsilon = 10$ was used to create adversarial examples. For SCGM, the maximum number of epochs was set to 300, and learning rate decreased by a factor of 10 each 100 epochs. All other parameters remained the same for this dataset. For all running experiments, we used the ramp-up strategy of Eq. (\ref{equ:scheduler}) to set hyper-parameters $\paramCot$ and $\paramDiv$. We set $\epochIni$ to $1$ for $\paramCot$ and $20$ for $\paramDiv$, since adversarial noise is meaningless if networks are not training enough. 
Moreover, we used $\epochEnd=50$ for both $\paramCot$ and $\paramDiv$. Last, we set the maximum hyper-parameter value $\paramMax$ to $0.5$ for $\paramCot$ and $0.05$ for $\paramDiv$. Note that all hyper-parameters of our method, as well as comparison baselines described below, were selected using grid search on the validation set.


We report the average performance of individual models, as well as the performance of combining the prediction of all models using a voting strategy. In preliminary experiments, we observed that soft-voting usually outperformed hard-voting and thus only considered this strategy. Our deep co-training method is compared against three popular approaches for semi-supervised learning: the Pseudo\,Label algorithm \cite{Leepseudolabel}, VAT \cite{miyato2018virtual} and Mean\,Teacher \cite{perone2018unsupervised}. To our knowledge, Mean\,Teacher is the only other approach using multiple deep CNNs for semi-supervised segmentation. For these three baselines, we follow the same optimization, learning rate decay, weight scheduler, and data augmentation setting as for our method. For the Pseudo\,Label algorithm, we consider the $\alpha\%$ most confident pixels of a prediction as ground truth, and increase $\alpha$ from $50\%$ to $99\%$ over training epochs. For VAT, we apply the same adversarial attack setting as in our method. For Mean\,Teacher, as in \cite{perone2018unsupervised}, data augmentation is applied to input images of a student model and Non-augmented images are fed to a teacher model, whose parameters $\theta'$ are computed by running an exponential moving average on the student's parameters $\theta$:
\beq
    \theta'_t \ = \ \alpha\theta'_{t-1} \, + \, (1-\alpha)\theta_t.
\eeq
In our experiments, we set $\alpha$ to 0.99. Finally, the student's output for augmented images is forced to be consistent with the teacher's prediction, augmented using the same strategy, via an $L_2$ loss. We then report the performance of the teacher network. 

\subsection{Experimental results}

\subsubsection{ACDC dataset}

We first evaluate our deep co-training method on the ACDC dataset using a dual view setting, i.e., training two segmentation models using the proposed loss. Performance is measured for individual models (we report their mean accuracy), as well as for the combined prediction using soft-voting. To simulate different levels of supervision, we vary the ratio $l_{a}$ of labeled images in the training set, $0 \leq l_{a} \leq 1$. Images and ground-truth segmentation masks from the first $75 \times l_{a}$ training subjects are used as labeled data, while the images of remaining subjects serve as unlabeled data.

As additional baseline for an ablation study, we trained the two models independently, 
without considering the ensemble agreement (i.e., $\LossCot$) or adversarial diversity (i.e., $\LossDiv$) loss terms. In the presentation of results, this baseline is referred to as Independent. Note that the soft-voting score of this baseline corresponds to the well-known bagging technique in ensemble learning. As fully-supervised baseline, we also report the performance obtained by training a single model with all available training examples. This baseline is denoted as full supervision (Full) in results. Moreover, to measure the relative contribution of the adversarial loss terms on performance, we also give the average and soft-voting score of the ensemble trained without this term, and denote this approach as JSD in the results. The proposed method, which combines all three loss terms, is referred to as Deep Co-Training Segmentation (DCT-Seg).

\begin{table}[t!]
\caption{DSC and HD performance of tested methods for validation images of the ACDC dataset. Except for full supervision (Full), all methods were trained with 20\% of labeled data. Independent \,($\LossSup$ only), JSD \,($\LossSup$+$\LossCot$) and DCT-Seg \,($\LossSup$+$\LossCot$+$\LossDiv$) were trained in a dual-view setting. For these methods, we report the average ensemble performance (avg) and the performance obtained by combining ensemble predictions with soft-voting (voting). \emph{Note}: reported values are the average (standard deviation in parenthesis) obtained over three separate runs, each one with a different random seed.
}
\label{tab:ACDC-compare}
\small
\centering
\setlength{\tabcolsep}{2.5pt}
\begin{tabular}{lcccccc}
\toprule
 \multicolumn{2}{c}{\multirow{2}{*}{\textbf{Method}}}  & & \multicolumn{4}{c}{\textbf{DSC ($\%$)}} \\ 
 \cmidrule(lr){4-7}
 & & & \textbf{RV} & \textbf{Myo}   & \textbf{LV}    & \textbf{Mean}  \\
\midrule
 \multicolumn{2}{l}{Full} & & 81.96\,(0.15) & 85.39\,(0.20) & 91.82\,(0.15) & 86.39\,(0.10) \\
 \midrule
 \multicolumn{2}{l}{Pseudo\,Label ~\cite{Leepseudolabel}} & & 74.60\,(0.32) & 78.91\,(0.21) & 85.79\,(0.17) & 79.77\,(0.14) \\ 
\multicolumn{2}{l}{VAT ~\cite{miyato2018virtual}} & & 72.78\,(0.39) & 80.81\,(0.21) & 87.60\,(0.18) & 80.39\,(0.15) \\ 
\multicolumn{2}{l}{Mean\,Teacher ~\cite{perone2018unsupervised}} & & 74.62\,(1.10) & 80.66\,(0.61)& 86.75\,(0.27) & 80.68\,(0.41) \\
 \midrule
 \multirow{2}{*}{Independent} & avg & & 68.82\,(1.90)  & 78.30\,(1.55) & 85.92\,(0.62) & 77.68\,(1.48) \\
 & voting & & 68.28\,(1.61) & 79.94\,(1.00) & 86.41\,(0.29) & 78.21\,(0.89) \\
\midrule
 \multirow{2}{*}{JSD} & avg & & 74.75\,(1.69) & 81.85\,(0.42) & 89.73\,(0.58) & 82.11\,(0.44) \\
 & voting & & 75.06\,(1.87) & 82.64\,(0.57) & \textbf{90.31\,(0.47)} & 82.67\,(0.67) \\
\midrule
 \multirow{2}{*}{DCT-Seg (ours)} & avg & & 77.51\,(0.69) &  82.43\,(0.27) & 89.85\,(0.26) & 83.26\,(0.16) \\
 & voting & & \textbf{78.20\,(0.70)} & \textbf{83.11\,(0.20)} & 90.22\,(0.24) & \textbf{83.84\,(0.10)} \\ 
\bottomrule
\end{tabular}

\vspace{5mm}

\setlength{\tabcolsep}{4.25pt}
\begin{tabular}{lcccccc}
\toprule
 \multicolumn{2}{c}{\multirow{2}{*}{\textbf{Method}}}  & & \multicolumn{4}{c}{\textbf{HD  (mm)}} \\ 
 \cmidrule(lr){4-7}
 & & & \textbf{RV} & \textbf{Myo}   & \textbf{LV}    & \textbf{Mean}  \\
\midrule\midrule
 \multicolumn{2}{l}{Full} & &11.42\,(1.15) & 5.80\,(0.98) & 4.58\,(0.65) & 7.27\,(0.50)\\
 \midrule
 \multicolumn{2}{l}{Pseudo\,Label ~\cite{Leepseudolabel}} & & 18.82\,(4.58) & 11.95\,(2.81)& 10.71\,(1.27) & 13.83\,(1.06)\\ 
\multicolumn{2}{l}{VAT ~\cite{miyato2018virtual}} & & 17.43\,(3.37) & 8.60\,(1.20)& 8.79\,(0.52) & 11.61\,(0.40)\\ 
\multicolumn{2}{l}{Mean\,Teacher ~\cite{perone2018unsupervised}} & & 16.12\,(1.12) & 7.86\,(0.78)& 7.41\,(0.57) & 10.46\,(0.36)  \\
\midrule
 \multirow{2}{*}{Independent} & avg & &
 21.26\,(3.04) & 13.31\,(2.17) & 9.21\,(1.95) & 14.59\,(1.87) \\
 & voting & & 8.77\,(2.18) & 6.65\,(1.83) & 5.00\,(1.13) & 6.81\,(0.72)\\
\midrule
 \multirow{2}{*}{JSD} & avg & & 15.84\,(1.59) & 7.18\,(0.47) & 5.63\,(0.88) & 9.55\,(0.30)\\
 & voting & & \textbf{7.39\,(0.77)} & 4.21\,(0.33) & \textbf{3.33\,(0.12)} & \textbf{4.97\,(0.04)}\\
\midrule
 \multirow{2}{*}{DCT-Seg (ours)} & avg & & 16.05\,(2.19) & 7.89\,(1.67) & 4.98\,(0.59) & 9.64\,(0.30)\\
 & voting & & 7.43\,(0.62) & \textbf{4.19\,(0.29)} & \textbf{3.33\,(0.09)} & 4.98\,(0.10) \\
\bottomrule
\end{tabular}
\end{table}

\begin{figure}[ht!]
 \centering
 \includegraphics[width=0.9\linewidth]{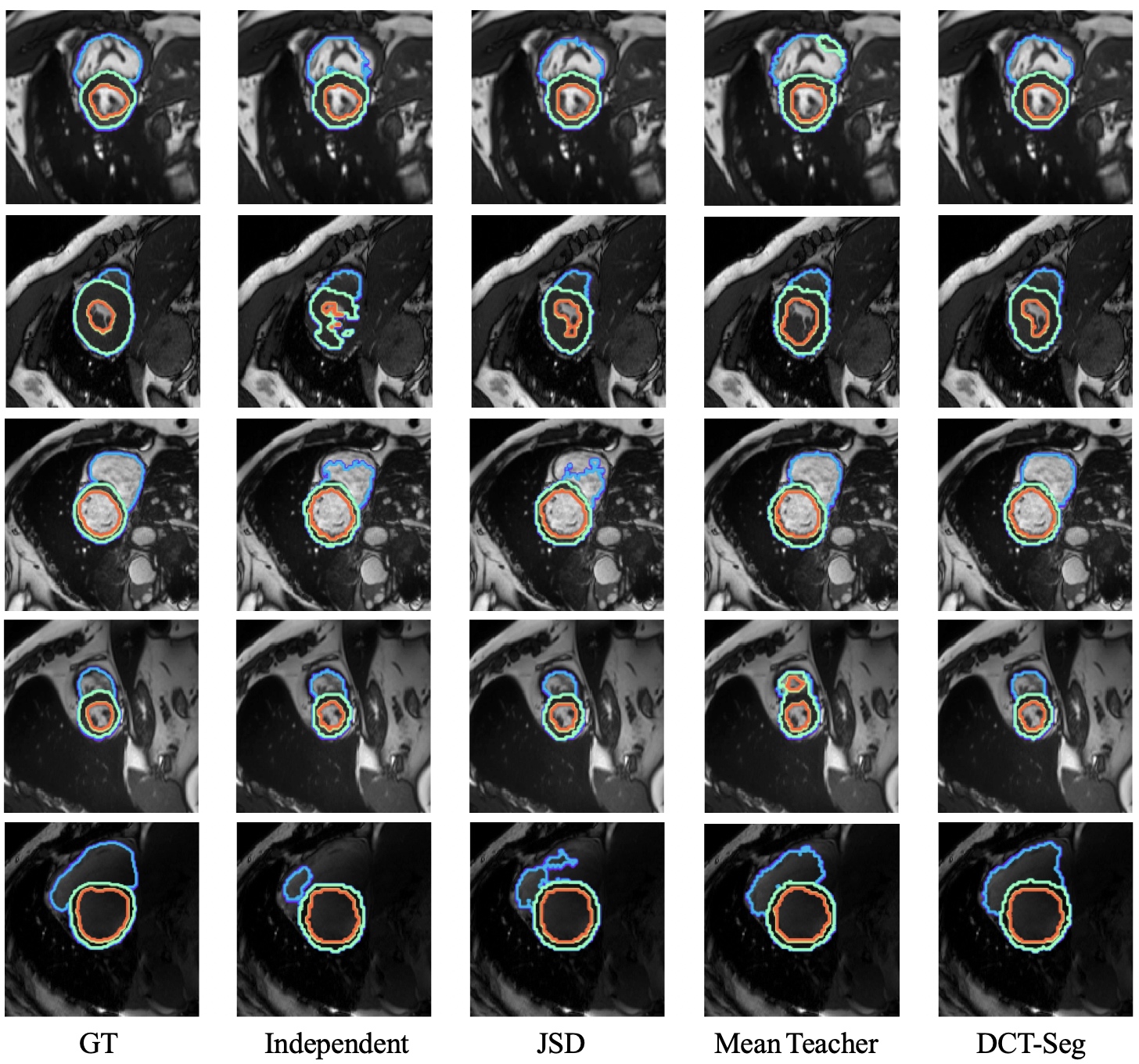}
 \caption{Examples of segmentation results for the ACDC dataset with 20\% of labeled training examples. From left to right: 
 Ground-truth (GT), Independent \,($\LossSup$ only), JSD \,($\LossSup$+$\LossCot$), Mean\,Teacher \cite{perone2018unsupervised}, and our DCT-Seg method \,($\LossSup$+$\LossCot$+$\LossDiv$). 
 }
 \label{fig:results_acdc}
\end{figure}

Table \ref{tab:ACDC-compare} gives the class-wise mean DSC and HD of tested methods for a labeled data ratio of $l_a = 0.2$. To evaluate robustness against parameter initialization, we ran the experiment three times with different random seeds, and computed the average and standard deviation of performance values over the three runs. We report both the ensemble average score (\emph{avg} in the table) and the score obtained by ensemble soft-voting (\emph{voting} in the table). 

For both DSC and HD, ensemble soft-voting leads to a higher accuracy than the prediction of individual models, in all cases. This confirms the benefit of aggregating predictions from different models. It can also be observed that considering ensemble agreement without diversity (JSD) leads to a higher accuracy than the supervised loss alone (Independent). For DSC, combining all three losses in DCT-Seg gives the best performance, with overall mean improvements of 5.63\% compared to Independent and 3.16\% over Mean\,Teacher. With only 20\% of training images labeled, DCT-Seg provides a mean DSC only 2.55\% less than full supervision. With respect to HD, our DCT-Seg method outperformed all three baselines by a significant margin. However, enforcing model diversity did not lead to noticeable improvements in this case, with DCT-Seg achieving a performance similar to JSD. This can potentially be explained by the fact that HD is more sensitive to outliers that can result from adversarial training. Examples of segmentation results for tested methods are shown in Fig. \ref{fig:results_acdc}. We see that deep co-training gives contours closer to the ground-truth, with very few artifacts on the boundaries between different regions.

\begin{table}[ht!]
\caption{DSC performance on the ACDC validation set when training different numbers of segmentation models (i.e., views) separately (Independent) or with the proposed deep co-training method (DCT-Seg). In this experiment, 20\% of training images are labeled. \emph{Note}: reported values are the average (standard deviation in parenthesis) obtained over three separate runs, each one with a different random seed.}
\label{tab:impact-views}
\centering
\begin{small}
\begin{tabular}{lcccc}
\toprule
\multicolumn{2}{c}{\textbf{Method}} & \textbf{2 views} & \textbf{3 views} & \textbf{4 views} \\
\midrule\midrule
 \multirow{2}{*}{Independent} & avg & 77.68\,(1.48)	& 77.80\,(1.27) & 77.82\,(0.92) \\
& voting & 78.21\,(0.89) & 78.57\,(0.71) & 79.08\,(1.21) \\
\midrule
\multirow{2}{*}{DCT-Seg (ours)} & avg & 83.26\,(0.16) & 83.80\,(0.38) & 83.43\,(0.24) \\
& voting & \textbf{83.84\,(0.10)} & \textbf{84.71\,(0.53)} & \textbf{84.61\,(0.28)}\\
\bottomrule
\end{tabular}
\end{small}
\end{table}

Next, we assess whether having more models in the ensemble (i.e., more than two views) can further boost performance of methods. Toward this goal, we repeated the experiment with 2, 3 and 4 views, once more using a labeled image ratio of $l_a = 0.2$. The overall mean DSC of tested methods, computed over the all classes, is reported in Table \ref{tab:impact-views}. We see that increasing the number of views does not significantly improve the performance for individually-trained models (Independent). On the other hand, for deep co-training, a small increase in DSC is observed when going from 2 to 3 views.  However, adding a fourth view does not further improve performance, suggesting that co-training can effectively capture variability with a very limited number of views.

\begin{table}[ht!]
\caption{DSC performance on the ACDC validation set when training two segmentation models separately (Independent) or with the proposed deep co-training method (DCT-Seg), for three different ratios $l_a$ of labeled examples. \emph{Note}: reported values are the average (standard deviation in parenthesis) obtained over three separate runs, each one with a different random seed.}
\label{tab:impact-labeled}
\centering
\small
\setlength{\tabcolsep}{2.5pt}
\begin{tabular}{lcccccc}
\toprule                
\multicolumn{2}{c}{\textbf{Method}} & & $\bm{l_a=5\%}$ & $\bm{l_a=10\%}$ & $\bm{l_a=20\%}$ & $\bm{l_a=50\%}$ \\
\midrule\midrule
 \multirow{2}{*}{Independent} & avg & & 69.72\,(0.10)&74.68\,(0.58) & 77.68\,(1.48) &84.96\,(0.13) \\
& voting & & 71.17\,(0.19)&75.84\,(0.49) & 78.21\,(0.89) & 85.12\,(0.08) \\
\midrule
\multirow{2}{*}{DCT-Seg (ours)} & avg & &77.81\,(0.10)& 82.36\,(0.33) & 83.26\,(0.16) & 	86.02\,(0.14) \\
& voting & & \textbf{78.17\,(0.12)} &\textbf{82.90\,(0.22)} & \textbf{83.84\,(0.10)} & \textbf{86.15\,(0.09)} \\
\bottomrule
\end{tabular}
\end{table}

As third experiment, we evaluate how the proportion of labeled data impacts results in a dual-view setting. Table \ref{tab:impact-labeled} gives the performance of individually-trained models (Independent) and co-training for three labeled data ratio: 10\%, 20\% and 50\%. A clear trend is observed in these results, where mean DSC values increase sharply with the ratio of labeled images in training. In all cases, deep co-training leads to a higher DSC than training models separately, the most significant improvements obtained for the smallest ratios of $l_a = 0.05$ ($7.00\%$) and $l_a = 0.1$ ($7.06\%$). 

\subsubsection{SCGM dataset}

To further validate the effectiveness of our proposed deep co-training method, we evaluated it on the task of segmenting spinal chord grey matter in images from the SCGM dataset. As mentioned previously, this experiment aims at testing our method in a challenging setting where very few labeled images are used in training (i.e., only 30 images), and test images are generated using different acquisition parameters.

\begin{figure}[ht!]
 \centering 
 \includegraphics[width=0.9\linewidth]{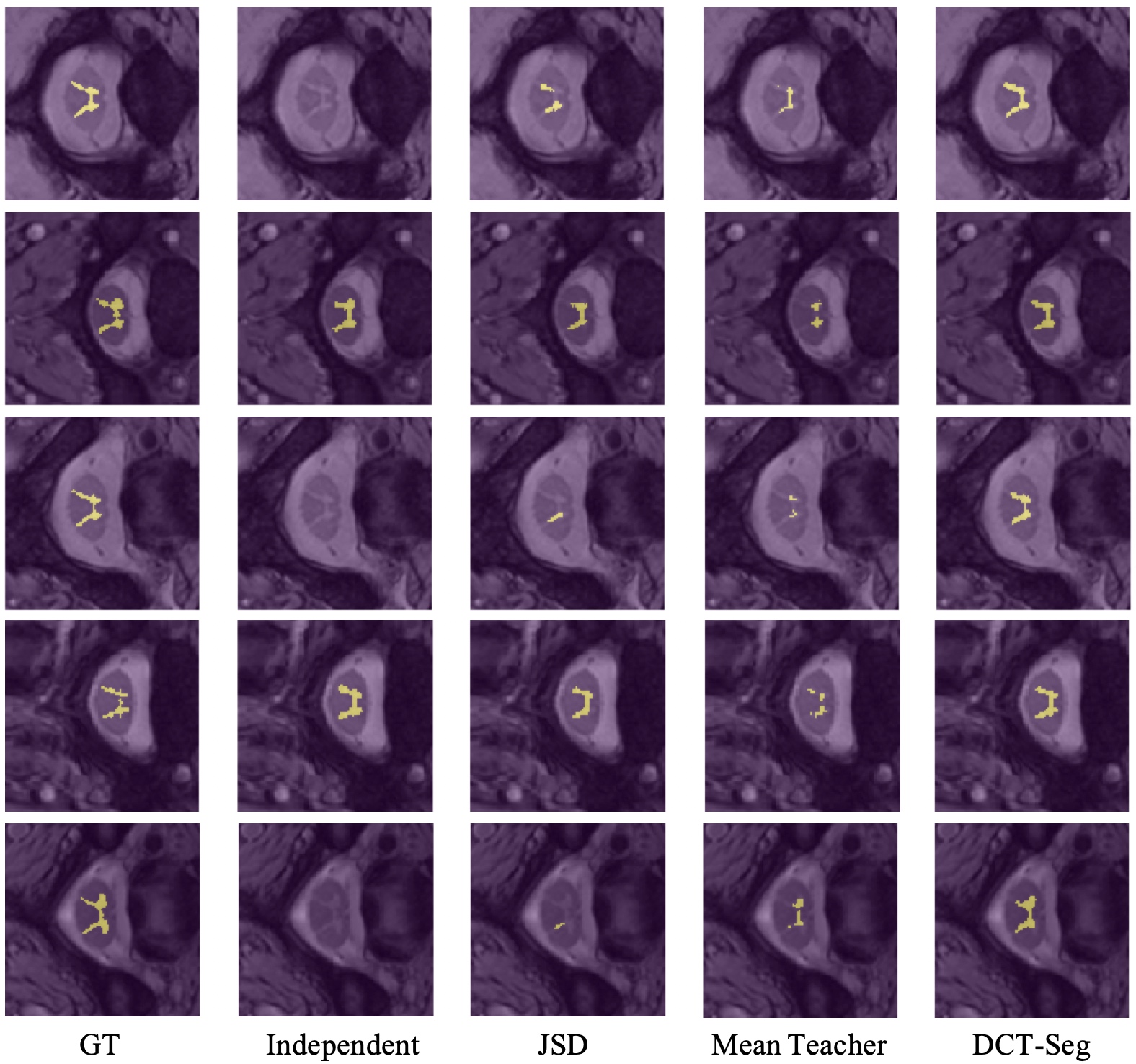}
 \caption{Examples of segmentation results for the SCGM dataset using Center 1 as training data. From left to right: Ground-truth (GT), Independent \,($\LossSup$ only), JSD \,($\LossSup$+$\LossCot$), Mean\,Teacher \cite{perone2018unsupervised}, and our DCT-Seg method \,($\LossSup$+$\LossCot$+$\LossDiv$). 
 }
 \label{fig:scgm-examples}
\end{figure}

\begin{table}[ht!]
\caption{
DSC performance of tested methods for validation images of the SCGM dataset. Independent ($\LossSup$ only), JSD ($\LossSup$+$\LossCot$) and DCT-Seg ($\LossSup$+$\LossCot$+$\LossDiv$) were trained in a dual-view setting. For these methods, we report the average ensemble performance and the DSC obtained by combining ensemble predictions with soft-voting. \emph{Note}: reported values are the average from two separate runs, each one with a different random seed.
}
\label{tab:SPGM-compare}
\small
\centering
\setlength{\tabcolsep}{6pt}
\begin{tabular}{lccc}
\toprule
 \multicolumn{2}{c}{\textbf{Method}}  & & \textbf{DSC}  \\
\midrule\midrule
   \multicolumn{2}{l}{Pseudo\,Label ~\cite{Leepseudolabel}} & & 60.03 \\
  \multicolumn{2}{l}{VAT ~\cite{miyato2018virtual}} & & 59.40 \\
 \multicolumn{2}{l}{Mean\,Teacher ~\cite{perone2018unsupervised}} & & 50.55 \\
 \midrule
 \multirow{2}{*}{Independent} & avg & & 43.31 \\
 & voting & & 43.22 \\
\midrule
 \multirow{2}{*}{JSD} & avg & & 45.59 \\
 & voting & & 44.96 \\
\midrule
 \multirow{2}{*}{DCT-Seg (ours)} & avg & & 71.09  \\
 & voting & & \textbf{72.76} \\
\bottomrule
\end{tabular}
\end{table}
Results of this experiment are summarized in Table \ref{tab:SPGM-compare}. Important differences can be observed between the DSC of tested methods. In this case, JSD improves the results of Independent only slightly, while deep co-training increases DSC scores of both these methods by nearly 25\%. This suggests that adversarial learning is highly useful when supervised training is limited (i.e., few labeled training examples, different from test examples). Compared to other tested semi-supervised approaches, our DCT-Seg method gives a mean DSC 12\% higher than the best baseline (Pseudo\,Label). The accuracy of deep co-training can be appreciated in Fig. \ref{fig:scgm-examples}, which shows examples of segmentation results for tested methods.

\begin{table}[ht!]
\caption{
DSC performance of tested methods for validation images of the Spleen dataset with resolutions of 256$\times$256 and 512$\times$512. Independent ($\LossSup$ only), JSD ($\LossSup$+$\LossCot$) and DCT-Seg ($\LossSup$+$\LossCot$+$\LossDiv$) were trained in a dual-view setting. For these methods, we report the average ensemble performance and the DSC obtained by combining ensemble predictions with soft-voting. \emph{Note}: reported values are the average from two separate runs, each one with a different random seed.
}
\label{tab:spleen-compare}
\small
\centering
\setlength{\tabcolsep}{6pt}
\begin{tabular}{lccc}
\hline
\toprule
\multirow{2}{*}{\textbf{Method}} &  & \multicolumn{2}{c}{\textbf{DSC (\%)}} \\
\cmidrule(lr){3-4}
 &  & 256$\times$256 & 512$\times$512 \\ 
 \midrule\midrule
Pseudo\,Label ~\cite{Leepseudolabel} &  & 85.71 & 84.83 \\
VAT ~\cite{miyato2018virtual} &  & 86.82 & 87.16 \\
Mean\,Teacher ~\cite{perone2018unsupervised} &  & 86.87 & 87.55 \\ \hline
\multirow{2}{*}{Independent} & avg & 84.71 & 86.63 \\
 & voting & 86.21 & 89.35 \\ \hline
\multirow{2}{*}{JSD} & avg & 87.92 & 90.04 \\
 & voting & 88.96 & 90.73 \\ \hline
\multirow{2}{*}{DCT-Seg (ours)} & avg & 89.30 & 91.06 \\
 & voting & \textbf{90.19} & \textbf{91.81} \\ 
 \bottomrule
\end{tabular}
\end{table}

\begin{figure}[ht!]
 \centering 
 \includegraphics[width=0.9\linewidth]{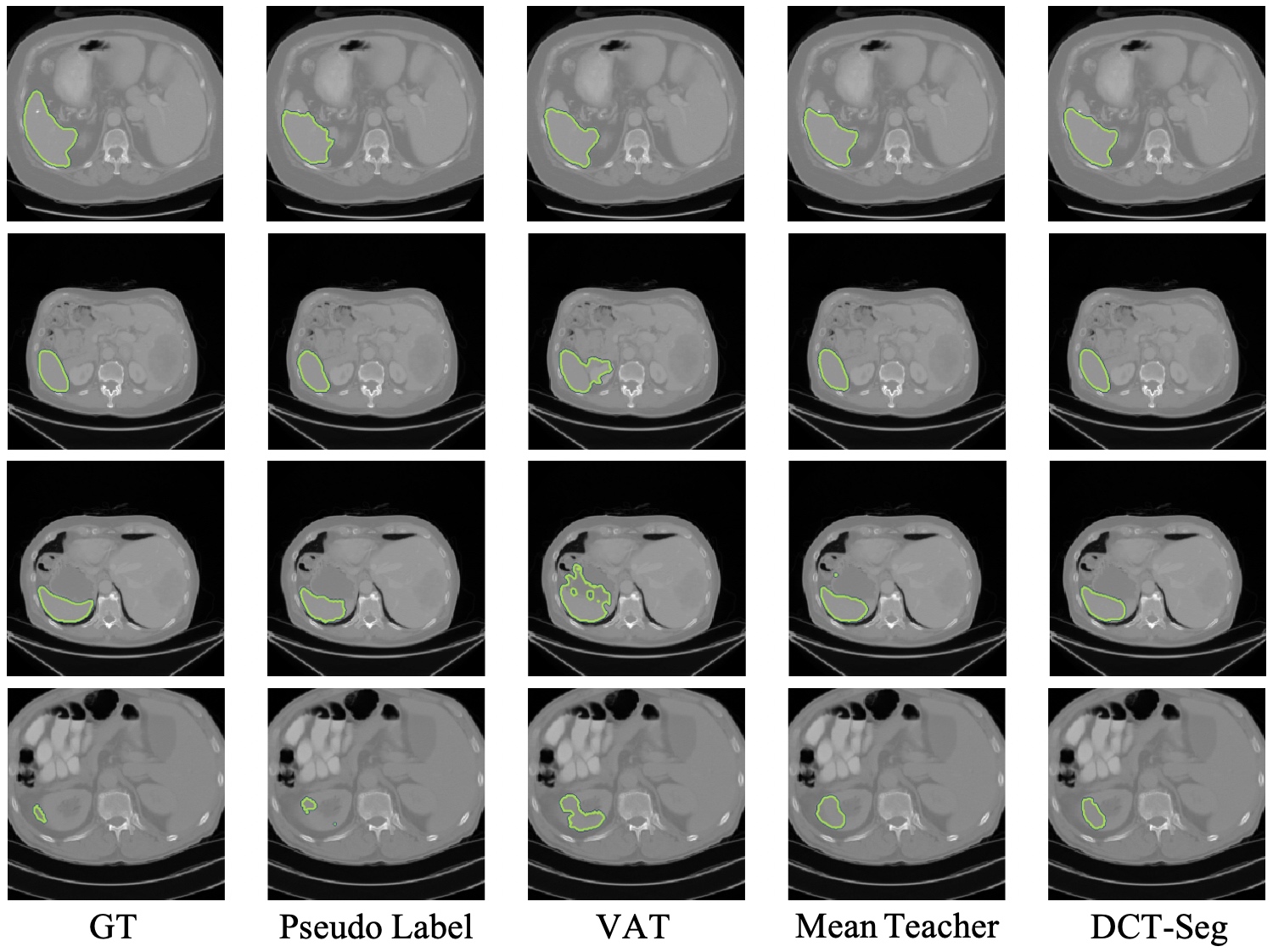}
 \caption{Examples of segmentation results of tested methods on the Spleen dataset with resolution of 256$\times$256. Note: our DCT-Seg method combines the predictions of two CNNs trained with the same subset of labeled examples as other approaches.
 }
 \label{fig:spleen-examples}
\end{figure}

\subsection{Spleen dataset}

We then investigate the robustness of our proposed algorithm to different data modalities and input resolutions. Toward this goal, we repeated our experiments on the Spleen dataset consisting of 2D slices of CT scans resized to a resolution of 256$\times$256 or 512$\times$512. 
Table \ref{tab:spleen-compare} summarizes the experimental results. We see that, regardless the input image size, our proposed method achieves a consistent improvement over other semi-supervised approaches. Specifically, the soft-voting version of DCT-Seg obtains a mean DSC boost of 3-4\% compared to the best performing baseline (Mean Teacher), showing its advantage for different image modalities and resolutions. Examples of segmentation results obtained by tested methods on images of size 256$\times$256 are given in Fig. \ref{fig:spleen-examples}. Visually, DCT-Seg and Mean Teacher provide similar results, with most pronounced differences observed for small foreground regions (e.g., last row of the figure).


\subsection{Impact of diversity loss}

We investigate the role of the ensemble diversity loss (i.e., $\LossDiv$) in our deep co-training method and experimentally show that it also acts as a coarse measure of model agreement, merging the prediction of models while avoiding them to collapse on each other. We perform our investigation on the ACDC dataset using two models. The first one is pre-trained by full supervision as a fixed reference, and the second one trained from scratch using a labeled data ratio of $l_a = 0.5$. Note that the trained model is only linked to the fixed reference by $\LossDiv$, and no supervised loss is considered while training this model. Moreover, to measure the impact of adversarial noise $\epsilon$ in $\LossDiv$, we repeat training with different values for $\epsilon$.  

\begin{figure}[ht!]
\centering
\includegraphics[width=\linewidth]{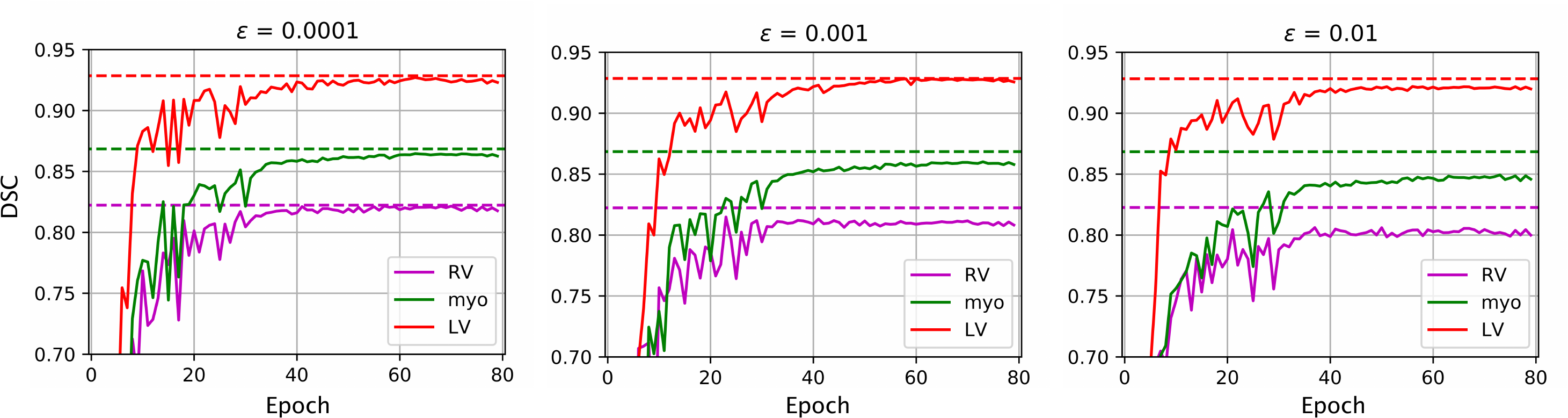}      
    \caption{DSC score for models trained from scratch using only $\LossDiv$ with different $\epsilon$. It can be seen that $\LossDiv$ acts as a similarity loss, especially when $\epsilon$ is small.}
    \label{fig:impact-epsilon-dif}
\end{figure}

Fig. \ref{fig:impact-epsilon-dif} gives the DSC obtained on the validation set by the reference model (dashed line) and model trained from scratch (solid line), for increasing amounts of adversarial noise $\epsilon$. It can be observed that the trained model rapidly converges to the reference, without the need for a supervised signal or specific agreement loss. However, upon convergence, we see that the trained model does not fully reach the accuracy of the reference model, and that the gap between the two models is proportional to the value of $\epsilon$. For example, a gap of 1.43\%, 1.38\% and 0.02\% is obtained for the Myo class, when using an $\epsilon$ of 0.01, 0.001, and 0.0001, respectively. This can be explained by the fact that, when $\epsilon$ is small, adversarial examples are very similar to original images, and $\LossDiv$ then acts as a symmetric KL loss between the two models.

\begin{figure}[ht!]
\centering
\includegraphics[width=1\linewidth]{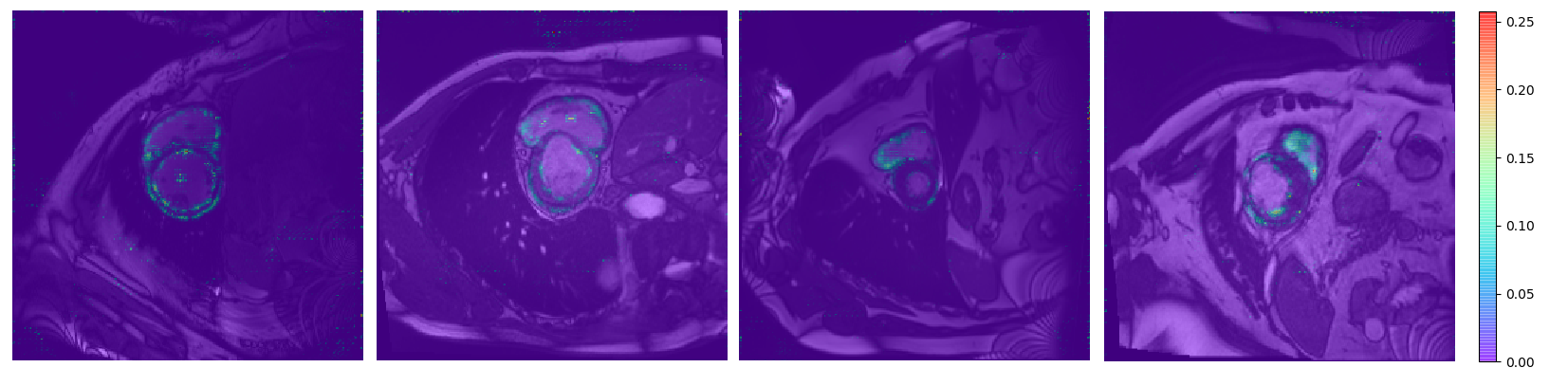}
\caption{
Examples of prediction disagreement between two models linked with the ensemble diversity loss ($\LossDiv$), measured using $L_1$ norm.
}
\label{fig:model-disagreement}
\end{figure}

We then tested the behavior of $\LossDiv$ when models are trained simultaneously. Toward this goal, we initialized the two models using the same fully-supervised checkpoint and linked them only using $\LossDiv$. Thus, the models give the same predictions at the beginning of training. As training progresses, $\LossDiv$ is minimized and the models should become different from one another. We show this tendency by imposing a small $\epsilon = 0.001$ during training. 
With the decrease of $\LossDiv$, differences start appearing along region boundaries, leading to slightly worse DSC scores. Examples of prediction disagreement, measured by the $L_1$ norm, are shown in Fig. \ref{fig:model-disagreement}. It can be observed that most prediction differences occur at the boundary and within regions which are hardest to segment (i.e., left ventricle myocardium and right ventricle endocardium).

\begin{figure}[ht!]
\centering
\includegraphics[width=.95\linewidth]{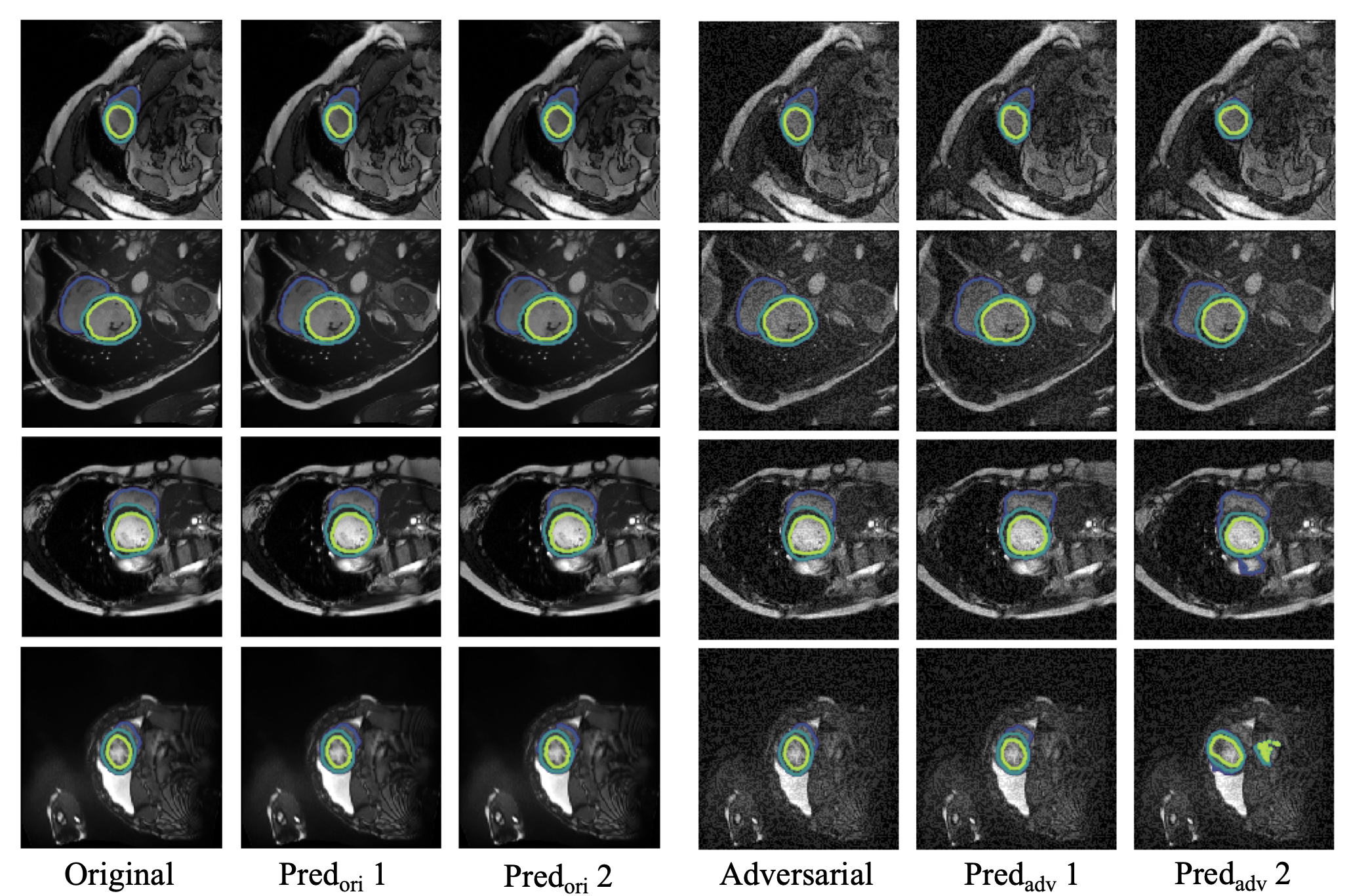}
\caption{
Impact of adversarial noise on prediction diversity. From left to right: original image (with GT contours), predictions of models 1 and 2 for the original image, adversarial image for model 2 (with GT contour), and predictions of model 1 and 2 for the adversarial image.
}
\label{fig:model-diversity}
\end{figure}

Last, we illustrate in Fig. \ref{fig:model-diversity} the effect of adversarial examples on model prediction diversity. For an input image, the two models can offer similar predictions. However, if this image is modified using adversarial noise, the predictions of the two models can differ significantly from one another. This confirms the usefulness of adversarial training for generating diversity between models.

\section{Discussion and conclusion}

We proposed the first application of deep co-training to single image segmentation and demonstrated its usefulness on two public benchmark datasets. Our experiments showed that both ensemble agreement and diversity loss terms help boost performance compared to standard techniques such as bagging, and that combining both in a deep co-training algorithm outperforms recent approaches like Pseudo\,Label, VAT and Mean\,Teacher.

A limitation of the proposed method is the need to train multiple segmentation networks at the same time, which increases the computational requirements and restricts the number of views possible. During testing, computing and combining multiple segmentation predictions also entails greater computational resources, although these predictions can be obtained in parallel (e.g., on separate GPUs). Nevertheless, our experiments on the ACDC dataset suggest that increasing the number of segmentation models beyond two offers limited benefits, showing the ability of our diversity-inducing strategy to capture variability in the data.

Another possible drawback of our method is the need to balance three different loss terms (i.e., $\LossSup$, $\LossCot$ and $\LossDiv$) that can compete against one another during training. To alleviate this problem, we proposed a ramp-up strategy where a greater importance is given to the supervised loss in initial training epochs. However, this strategy still requires some tuning which can affect performance. A useful extension of this work could be to investigate self-tuning mechanisms which can adapt more efficiently to new datasets.

In this work, an adversarial learning technique was employed to enforce diversity in the ensemble models. As shown in our results, this technique can also push the predictions of models toward each other, and generates differences mostly at the boundary or within hard-to-segment regions. As future work, it would be interesting to explore a broader range of strategies to create diversity, for example using fake images from generative adversarial networks. Moreover, our experiments revolved around three different medical image segmentation problems and included images from both MRI and CT modalities. As motivated in the introduction, semi-supervised learning is most important for medical applications, where annotating images is complex and expensive. Nonetheless, evaluating the proposed method on additional types of images and segmentation tasks would help to further validate its usefulness.
         
\section*{Acknowledgements}  

We thank NVIDIA corporation for supporting this work through their GPU grant program. This project is partially supported by FRQNT scholarship.

\section*{References}

\bibliography{mybibfile}

\begin{thebibliography}{10}
\expandafter\ifx\csname url\endcsname\relax
  \def\url#1{\texttt{#1}}\fi
\expandafter\ifx\csname urlprefix\endcsname\relax\def\urlprefix{URL }\fi
\expandafter\ifx\csname href\endcsname\relax
  \def\href#1#2{#2} \def\path#1{#1}\fi

\bibitem{noh2015learning}
H.~Noh, S.~Hong, B.~Han, Learning deconvolution network for semantic
  segmentation, in: Proceedings of the IEEE international conference on
  computer vision, 2015, pp. 1520--1528.

\bibitem{Litjens2017}
G.~J.~S. Litjens, T.~Kooi, B.~E. Bejnordi, A.~A.~A. Setio, F.~Ciompi,
  M.~Ghafoorian, J.~A. W.~M. van~der Laak, B.~van Ginneken, C.~I.
  S{\'{a}}nchez, A survey on deep learning in medical image analysis, Medical
  Image Analysis 42 (2017) 60--88.
\newblock \href {http://dx.doi.org/10.1016/j.media.2017.07.005}
  {\path{doi:10.1016/j.media.2017.07.005}}.

\bibitem{long2015fully}
J.~Long, E.~Shelhamer, T.~Darrell, Fully convolutional networks for semantic
  segmentation, in: Proceedings of the IEEE conference on computer vision and
  pattern recognition, 2015, pp. 3431--3440.

\bibitem{milletari2016v}
F.~Milletari, N.~Navab, S.-A. Ahmadi, V-net: Fully convolutional neural
  networks for volumetric medical image segmentation, in: 3D Vision (3DV), 2016
  Fourth International Conference on, IEEE, 2016, pp. 565--571.

\bibitem{kolesnikov2016seed}
A.~Kolesnikov, C.~H. Lampert, Seed, expand and constrain: Three principles for
  weakly-supervised image segmentation, in: European Conference on Computer
  Vision, Springer, 2016, pp. 695--711.

\bibitem{wang2019benchmark}
L.~Wang, D.~Nie, G.~Li, {\'E}.~Puybareau, J.~Dolz, Q.~Zhang, F.~Wang, J.~Xia,
  Z.~Wu, J.~Chen, et~al., Benchmark on automatic 6-month-old infant brain
  segmentation algorithms: The iseg-2017 challenge, IEEE transactions on
  medical imaging.

\bibitem{pinheiro2015weakly}
P.~O. Pinheiro, R.~Collobert, Weakly supervised semantic segmentation with
  convolutional networks, in: CVPR, Vol.~2, Citeseer, 2015, p.~6.

\bibitem{papandreou2015weakly}
G.~Papandreou, L.-C. Chen, K.~Murphy, A.~L. Yuille, Weakly-and semi-supervised
  learning of a deep {CNN} for semantic image segmentation, arXiv preprint
  arXiv:1502.02734.

\bibitem{kervadec2019constrained}
H.~Kervadec, J.~Dolz, M.~Tang, E.~Granger, Y.~Boykov, I.~B. Ayed,
  Constrained-{CNN} losses for weakly supervised segmentation, Medical image
  analysis.

\bibitem{pathak2015constrained}
D.~Pathak, P.~Krahenbuhl, T.~Darrell, Constrained convolutional neural networks
  for weakly supervised segmentation, in: Proceedings of the IEEE international
  conference on computer vision, 2015, pp. 1796--1804.

\bibitem{dai2015boxsup}
J.~Dai, K.~He, J.~Sun, Boxsup: Exploiting bounding boxes to supervise
  convolutional networks for semantic segmentation, in: Proceedings of the IEEE
  International Conference on Computer Vision, 2015, pp. 1635--1643.

\bibitem{rajchl2017deepcut}
M.~Rajchl, M.~C. Lee, O.~Oktay, K.~Kamnitsas, J.~Passerat-Palmbach, W.~Bai,
  M.~Damodaram, M.~A. Rutherford, J.~V. Hajnal, B.~Kainz, et~al., Deepcut:
  Object segmentation from bounding box annotations using convolutional neural
  networks, IEEE transactions on medical imaging 36~(2) (2017) 674--683.

\bibitem{lin2016scribblesup}
D.~Lin, J.~Dai, J.~Jia, K.~He, J.~Sun, Scribblesup: Scribble-supervised
  convolutional networks for semantic segmentation, in: Proceedings of the IEEE
  Conference on Computer Vision and Pattern Recognition, 2016, pp. 3159--3167.

\bibitem{vezhnevets2010towards}
A.~Vezhnevets, J.~M. Buhmann, Towards weakly supervised semantic segmentation
  by means of multiple instance and multitask learning, in: Proceedings of the
  IEEE Conference on Computer Vision and Pattern Recognition, IEEE, 2010, pp.
  3249--3256.

\bibitem{bearman2016s}
A.~Bearman, O.~Russakovsky, V.~Ferrari, L.~Fei-Fei, What’s the point:
  Semantic segmentation with point supervision, in: European Conference on
  Computer Vision, Springer, 2016, pp. 549--565.

\bibitem{wei2016learning}
Y.~Wei, X.~Liang, Y.~Chen, Z.~Jie, Y.~Xiao, Y.~Zhao, S.~Yan, Learning to
  segment with image-level annotations, Pattern Recognition 59 (2016) 234--244.

\bibitem{pinheiro2015image}
P.~O. Pinheiro, R.~Collobert, From image-level to pixel-level labeling with
  convolutional networks, in: Proceedings of the IEEE Conference on Computer
  Vision and Pattern Recognition, 2015, pp. 1713--1721.

\bibitem{qi2016augmented}
X.~Qi, Z.~Liu, J.~Shi, H.~Zhao, J.~Jia, Augmented feedback in semantic
  segmentation under image level supervision, in: European Conference on
  Computer Vision, Springer, 2016, pp. 90--105.

\bibitem{saleh2016built}
F.~Saleh, M.~S. Aliakbarian, M.~Salzmann, L.~Petersson, S.~Gould, J.~M.
  Alvarez, Built-in foreground/background prior for weakly-supervised semantic
  segmentation, in: European Conference on Computer Vision, Springer, 2016, pp.
  413--432.

\bibitem{shimoda2016distinct}
W.~Shimoda, K.~Yanai, Distinct class-specific saliency maps for weakly
  supervised semantic segmentation, in: European Conference on Computer Vision,
  Springer, 2016, pp. 218--234.

\bibitem{hou2017deeply}
Q.~Hou, M.-M. Cheng, X.~Hu, A.~Borji, Z.~Tu, P.~Torr, Deeply supervised salient
  object detection with short connections, in: 2017 IEEE Conference on Computer
  Vision and Pattern Recognition (CVPR), IEEE, 2017, pp. 5300--5309.

\bibitem{liu2016dhsnet}
N.~Liu, J.~Han, Dhsnet: Deep hierarchical saliency network for salient object
  detection, in: Proceedings of the IEEE Conference on Computer Vision and
  Pattern Recognition, 2016, pp. 678--686.

\bibitem{selvaraju2017grad}
R.~R. Selvaraju, M.~Cogswell, A.~Das, R.~Vedantam, D.~Parikh, D.~Batra, et~al.,
  Grad-cam: Visual explanations from deep networks via gradient-based
  localization., in: ICCV, 2017, pp. 618--626.

\bibitem{zhou2016learning}
B.~Zhou, A.~Khosla, A.~Lapedriza, A.~Oliva, A.~Torralba, Learning deep features
  for discriminative localization, in: Proceedings of the IEEE Conference on
  Computer Vision and Pattern Recognition, 2016, pp. 2921--2929.

\bibitem{bai2017semi}
W.~Bai, O.~Oktay, M.~Sinclair, H.~Suzuki, M.~Rajchl, G.~Tarroni, B.~Glocker,
  A.~King, P.~M. Matthews, D.~Rueckert, Semi-supervised learning for
  network-based cardiac mr image segmentation, in: International Conference on
  Medical Image Computing and Computer-Assisted Intervention, Springer, 2017,
  pp. 253--260.

\bibitem{baur2017semi}
C.~Baur, S.~Albarqouni, N.~Navab, Semi-supervised deep learning for fully
  convolutional networks, in: International Conference on Medical Image
  Computing and Computer-Assisted Intervention, Springer, 2017, pp. 311--319.

\bibitem{min2018robust}
S.~Min, X.~Chen, A robust deep attention network to noisy labels in
  semi-supervised biomedical segmentation, arXiv preprint arXiv:1807.11719.

\bibitem{zhou2018semi}
Y.~Zhou, Y.~Wang, P.~Tang, W.~Shen, E.~K. Fishman, A.~L. Yuille,
  Semi-supervised multi-organ segmentation via multi-planar co-training, arXiv
  preprint arXiv:1804.02586.

\bibitem{perone2018unsupervised}
C.~S. Perone, P.~Ballester, R.~C. Barros, J.~Cohen-Adad, Unsupervised domain
  adaptation for medical imaging segmentation with self-ensembling, arXiv
  preprint arXiv:1811.06042.

\bibitem{gupta2016cross}
S.~Gupta, J.~Hoffman, J.~Malik, Cross modal distillation for supervision
  transfer, in: Proceedings of the IEEE Conference on Computer Vision and
  Pattern Recognition, 2016, pp. 2827--2836.

\bibitem{radosavovic2017data}
I.~Radosavovic, P.~Doll{\'a}r, R.~Girshick, G.~Gkioxari, K.~He, Data
  distillation: Towards omni-supervised learning, arXiv preprint
  arXiv:1712.04440.

\bibitem{souly2017semi}
N.~Souly, C.~Spampinato, M.~Shah, Semi supervised semantic segmentation using
  generative adversarial network, in: Computer Vision (ICCV), 2017 IEEE
  International Conference on, IEEE, 2017, pp. 5689--5697.

\bibitem{Hung_semiseg_2018}
W.-C. Hung, Y.-H. Tsai, Y.-T. Liou, Y.-Y. Lin, M.-H. Yang, Adversarial learning
  for semi-supervised semantic segmentation, in: Proceedings of the British
  Machine Vision Conference (BMVC), 2018, p.~1.

\bibitem{zhang2017deep}
Y.~Zhang, L.~Yang, J.~Chen, M.~Fredericksen, D.~P. Hughes, D.~Z. Chen, Deep
  adversarial networks for biomedical image segmentation utilizing unannotated
  images, in: International Conference on Medical Image Computing and
  Computer-Assisted Intervention, Springer, 2017, pp. 408--416.

\bibitem{luc2016semantic}
P.~Luc, C.~Couprie, S.~Chintala, J.~Verbeek, Semantic segmentation using
  adversarial networks, arXiv preprint arXiv:1611.08408.

\bibitem{blum1998combining}
A.~Blum, T.~Mitchell, Combining labeled and unlabeled data with co-training,
  in: Proceedings of the eleventh annual conference on Computational learning
  theory, ACM, 1998, pp. 92--100.

\bibitem{xu2013survey}
C.~Xu, D.~Tao, C.~Xu, A survey on multi-view learning, arXiv preprint
  arXiv:1304.5634.

\bibitem{wan2009co}
X.~Wan, Co-training for cross-lingual sentiment classification, in: Proceedings
  of the Joint Conference of the 47th Annual Meeting of the ACL and the 4th
  International Joint Conference on Natural Language Processing of the AFNLP:
  Volume 1-volume 1, Association for Computational Linguistics, 2009, pp.
  235--243.

\bibitem{nigam2000understanding}
K.~Nigam, R.~Ghani, Understanding the behavior of co-training, in: Proceedings
  of KDD-2000 workshop on text mining, Citeseer, 2000, pp. 15--17.

\bibitem{maeireizo2004co}
B.~Maeireizo, D.~Litman, R.~Hwa, Co-training for predicting emotions with
  spoken dialogue data, in: Proceedings of the ACL 2004 on Interactive poster
  and demonstration sessions, Association for Computational Linguistics, 2004,
  p.~28.

\bibitem{levin2003unsupervised}
A.~Levin, P.~A. Viola, Y.~Freund, Unsupervised improvement of visual detectors
  using co-training., in: ICCV, Vol.~1, 2003, pp. 626--633.

\bibitem{qiao2018deep}
S.~Qiao, W.~Shen, Z.~Zhang, B.~Wang, A.~Yuille, Deep co-training for
  semi-supervised image recognition, arXiv preprint arXiv:1803.05984.

\bibitem{Cooper1970On}
D.~Cooper, J.~Freeman, On the asymptotic improvement in the outcome of
  supervised learning provided by additional nonsupervised learning, IEEE
  Transactions on Computers 19~(11) (1970) 1055--1063.
\newblock \href {http://dx.doi.org/10.1109/T-C.1970.222832}
  {\path{doi:10.1109/T-C.1970.222832}}.

\bibitem{Dempster1977Maximum}
A.~P. Dempster, N.~M. Laird, D.~B. Rubin, Maximum likelihood from incomplete
  data via the em algorithm, Journal of the Royal Statistical Society, Series B
  39~(1) (1977) 1--38.

\bibitem{Chapelle2010Semi}
O.~Chapelle, B.~Schlkopf, A.~Zien, Semi-Supervised Learning, 1st Edition, The
  MIT Press, 2010.

\bibitem{Leepseudolabel}
D.~hyun Lee, Pseudo-label: The simple and efficient semi-supervised learning
  method for deep neural networks.

\bibitem{Grandvalet+Bengio-ssl-2006}
Y.~Grandvalet, Y.~Bengio, Entropy regularization, in: O.~Chapelle,
  B.~Sch{\"{o}}lkopf, A.~Zien (Eds.), Semi-Supervised Learning, {MIT} Press,
  2006, pp. 151--168.

\bibitem{Ramus15Semi}
A.~Rasmus, M.~Berglund, M.~Honkala, H.~Valpola, T.~Raiko, Semi-supervised
  learning with ladder networks, in: C.~Cortes, N.~D. Lawrence, D.~D. Lee,
  M.~Sugiyama, R.~Garnett (Eds.), Advances in Neural Information Processing
  Systems 28, Curran Associates, Inc., 2015, pp. 3546--3554.

\bibitem{Kingma14Semi}
D.~P. Kingma, S.~Mohamed, D.~Jimenez~Rezende, M.~Welling, Semi-supervised
  learning with deep generative models, in: Z.~Ghahramani, M.~Welling,
  C.~Cortes, N.~D. Lawrence, K.~Q. Weinberger (Eds.), Advances in Neural
  Information Processing Systems 27, Curran Associates, Inc., 2014, pp.
  3581--3589.

\bibitem{Laine16Temporal}
S.~Laine, T.~Aila, Temporal ensembling for semi-supervised learning, CoRR
  abs/1610.02242.
\newblock \href {http://arxiv.org/abs/1610.02242} {\path{arXiv:1610.02242}}.

\bibitem{Tarvainen17Weight}
A.~Tarvainen, H.~Valpola, Weight-averaged consistency targets improve
  semi-supervised deep learning results, CoRR abs/1703.01780.
\newblock \href {http://arxiv.org/abs/1703.01780} {\path{arXiv:1703.01780}}.

\bibitem{Goodfellow15Explaining}
I.~Goodfellow, J.~Shlens, C.~Szegedy, Explaining and harnessing adversarial
  examples, in: International Conference on Learning Representations, 2015,
  p.~1.

\bibitem{Miyato15Virtual}
T.~Miyato, S.~ichi Maeda, M.~Koyama, K.~Nakae, S.~Ishii, Distributional
  smoothing by virtual adversarial examples., CoRR abs/1507.00677.

\bibitem{Oliver2018Realistic}
A.~Oliver, A.~Odena, C.~A. Raffel, E.~D. Cubuk, I.~Goodfellow, Realistic
  evaluation of deep semi-supervised learning algorithms, in: S.~Bengio,
  H.~Wallach, H.~Larochelle, K.~Grauman, N.~Cesa-Bianchi, R.~Garnett (Eds.),
  Advances in Neural Information Processing Systems 31, Curran Associates,
  Inc., 2018, pp. 3239--3250.

\bibitem{Goodfellow2014GAN}
I.~Goodfellow, J.~Pouget-Abadie, M.~Mirza, B.~Xu, D.~Warde-Farley, S.~Ozair,
  A.~Courville, Y.~Bengio, Generative adversarial nets, in: Z.~Ghahramani,
  M.~Welling, C.~Cortes, N.~D. Lawrence, K.~Q. Weinberger (Eds.), Advances in
  Neural Information Processing Systems 27, Curran Associates, Inc., 2014, pp.
  2672--2680.

\bibitem{litjens2017survey}
G.~Litjens, T.~Kooi, B.~E. Bejnordi, A.~A.~A. Setio, F.~Ciompi, M.~Ghafoorian,
  J.~A. van~der Laak, B.~Van~Ginneken, C.~I. S{\'a}nchez, A survey on deep
  learning in medical image analysis, Medical image analysis 42 (2017) 60--88.

\bibitem{dolz20173d}
J.~Dolz, C.~Desrosiers, I.~Ben~Ayed, 3{D} fully convolutional networks for
  subcortical segmentation in {MRI}: {A} large-scale study, NeuroImage 170
  (2018) 456--470.

\bibitem{dolz2018hyperdense}
J.~Dolz, K.~Gopinath, J.~Yuan, H.~Lombaert, C.~Desrosiers, I.~B. Ayed,
  Hyperdense-net: A hyper-densely connected cnn for multi-modal image
  segmentation, IEEE transactions on medical imaging.

\bibitem{miyato2018virtual}
T.~Miyato, S.-i. Maeda, S.~Ishii, M.~Koyama, Virtual adversarial training: a
  regularization method for supervised and semi-supervised learning, IEEE
  transactions on pattern analysis and machine intelligence.

\bibitem{ACDC}
O.~B. et~al., Deep learning techniques for automatic {MRI} cardiac
  multi-structures segmentation and diagnosis: Is the problem solved?, IEEE
  Transactions on Medical Imaging 37~(11) (2018) 2514--2525.
\newblock \href {http://dx.doi.org/10.1109/TMI.2018.2837502}
  {\path{doi:10.1109/TMI.2018.2837502}}.

\bibitem{SCGM}
F.~Prados, J.~Ashburner, C.~Blaiotta, T.~Brosch, J.~Carballido-Gamio, M.~J.
  Cardoso, B.~N. Conrad, E.~Datta, G.~D{\'a}vid, B.~De~Leener, et~al., Spinal
  cord grey matter segmentation challenge, Neuroimage 152 (2017) 312--329.

\bibitem{simpson2019large}
A.~L. Simpson, M.~Antonelli, S.~Bakas, M.~Bilello, K.~Farahani, B.~van
  Ginneken, A.~Kopp-Schneider, B.~A. Landman, G.~Litjens, B.~Menze, et~al., A
  large annotated medical image dataset for the development and evaluation of
  segmentation algorithms, arXiv preprint arXiv:1902.09063.

\bibitem{ronneberger2015u}
O.~Ronneberger, P.~Fischer, T.~Brox, U-net: Convolutional networks for
  biomedical image segmentation, in: International Conference on Medical image
  computing and computer-assisted intervention, Springer, 2015, pp. 234--241.

\end{thebibliography}

\end{document}